\documentclass[conference]{IEEEtran}
\IEEEoverridecommandlockouts
\usepackage{cite}
\usepackage{amsmath,amssymb,amsfonts}
\usepackage{graphicx}
\usepackage{textcomp}
\usepackage{xcolor}

\usepackage{tcolorbox}

\usepackage{supertabular}
\usepackage{threeparttable}
\usepackage{soul}
\usepackage[ruled,linesnumbered]{algorithm2e}
\usepackage{colortbl}
\usepackage{listings}
\usepackage{multirow}
\usepackage{algpseudocode}
\usepackage{color}
\usepackage{moreverb}
\usepackage{mathrsfs}
\usepackage{bm}
\usepackage{framed}
\usepackage{bbding}
\usepackage{subfigure}
\usepackage[T1]{fontenc}
\usepackage{multicol}
\usepackage{enumitem}
\usepackage{balance}
\usepackage{url}

\newcommand{\qod}{\nobreak \ifvmode \relax \else
      \ifdim\lastskip<1.5em \hskip-\lastskip
      \hskip1.5em plus0em minus0.5em \fi \nobreak
      \vrule height0.5em width0.5em depth0em\fi}

\newtheorem{theorem}{Theorem}[section]
\newtheorem{lemma}[theorem]{Lemma}

\newtheorem{assumption}[theorem]{Assumption}

\newenvironment{proof}{{\noindent\it Proof:}\quad}{\hfill $\square$\par}

\def\BibTeX{{\rm B\kern-.05em{\sc i\kern-.025em b}\kern-.08em
    T\kern-.1667em\lower.7ex\hbox{E}\kern-.125emX}}
\begin{document}

\balance

\newpage

\title{FedCross: Towards Accurate Federated Learning via Multi-Model Cross-Aggregation
}

\author{
\IEEEauthorblockN{Ming Hu$^1$, Peiheng Zhou$^2$, Zhihao Yue$^2$, Zhiwei Ling$^2$, Yihao Huang$^1$, Anran Li$^1$, \\Yang Liu$^1$,  Xiang Lian$^3$, Mingsong Chen$^2$}
\IEEEauthorblockA{\textit{$^1$School of Computer Science and Engineering, Nanyang Technological University, Singapore} \\
\textit{$^2$MoE Engineering
Research Center of SW/HW Co-Design Tech. and App., East China Normal University, China}
\\
\textit{$^3$Department of Computer Science, Kent State University, Ohio, USA}
}
}

\maketitle

\begin{abstract}

As a promising distributed machine learning paradigm, Federated Learning (FL) has attracted increasing attention to deal with data silo problems without compromising user privacy. By adopting the classic one-to-multi training scheme (i.e., FedAvg), where the cloud server dispatches one single global model to multiple involved clients, conventional FL methods can achieve collaborative model training without data sharing. However, since only one global model cannot always accommodate all the incompatible convergence directions of local models, existing FL approaches greatly suffer from inferior classification accuracy.
To address this issue, we present an efficient FL framework named \textbf{\emph{FedCross}}, which uses a novel multi-to-multi FL training scheme based on our proposed \textbf{\emph{ multi-model cross-aggregation}} approach. Unlike traditional FL methods,  in each round of FL training, FedCross uses multiple middleware models to conduct weighted fusion individually. Since the middleware models used by FedCross can quickly converge into the same flat valley in terms of loss landscapes, the generated global model can achieve a well-generalization. Experimental results on various well-known datasets show that, compared with state-of-the-art FL methods, FedCross can significantly improve FL accuracy within both IID and non-IID scenarios without causing additional communication overhead.
\end{abstract}


\section{Introduction}

Along with the prosperity of Artificial Intelligence (AI) and Internet of Things (IoT)  techniques, more and more Artificial Intelligence of Things (AIoT) applications \cite{wang2023federated} (e.g., autonomous driving \cite{kdd2021}, smart transportation \cite{kdd2019},  medical monitoring \cite{iccad2020}) resort to Deep Neural Network (DNN) models to enable accurate sensing and intelligent control. Although such DNN models can deal with various complex tasks, due to the limited learning capabilities of IoT devices and stringent requirements for their data privacy, traditional centralized DNN training methods suffer a lot from the problem of low classification performance.
Alternatively, to facilitate the design of large-scale AIoT applications, Federated Learning (FL)~\cite{mcmahan2017communication,yang2019federated,zhang2020efficient,hu2023aiotml,li2021efficient} has been used as a promising distributed machine learning-based infrastructure, which allows knowledge sharing among AIoT devices without compromising their privacy.
Typically, FL adopts a cloud-client architecture, where the cloud server periodically updates the global model by aggregating the received local gradients and dispatching the updated global model to clients for a new round of training. Since none of the clients send their raw data to the cloud server, their privacy can be safely preserved.

Although FL is good at knowledge sharing among clients, it often fails to withstand low classification performance in deploying real-world applications, especially when client data are non-IID (Independent and Identically Distributed) \cite{wang2020optimizing,li2022federated,luo2021no,li2021fedrs, li2021sample}. 
This is mainly because most existing FL methods rely on the classic aggregation scheme, i.e., Federated Averaging (FedAvg)~\cite{mcmahan2017communication}, where the cloud server only dispatches one single global model to selected clients in a one-to-multi manner. 
Since the raw data on clients are different, the optimization directions of local models will gradually become divergent during the training,  resulting in conflicting gradients among local models.
In this case, by simply averaging the collected gradients from all the selected clients, the knowledge and efforts of local models accumulated in previous rounds of FL training are inevitably eclipsed. Due to such notorious phenomenon of gradient divergence~\cite{kairouz2021advances,zhuang2020performance}, the classification capability of the global model is greatly limited.
To alleviate the gradient divergence problem, 
 various approaches have been investigated to guide the optimization directions
 of local training, striving
 to derive local models with fewer conflicting parameters. 
However, since such methods cannot prevent the knowledge learned by individual clients from being damaged by the coarse-grained aggregation strategy (i.e.,  FedAvg), the classification capability of the global model is still restricted.
%

\begin{figure}[h] 
  \vspace{ -0.1 in}
	\begin{center} 
		\includegraphics[width=0.41\textwidth]{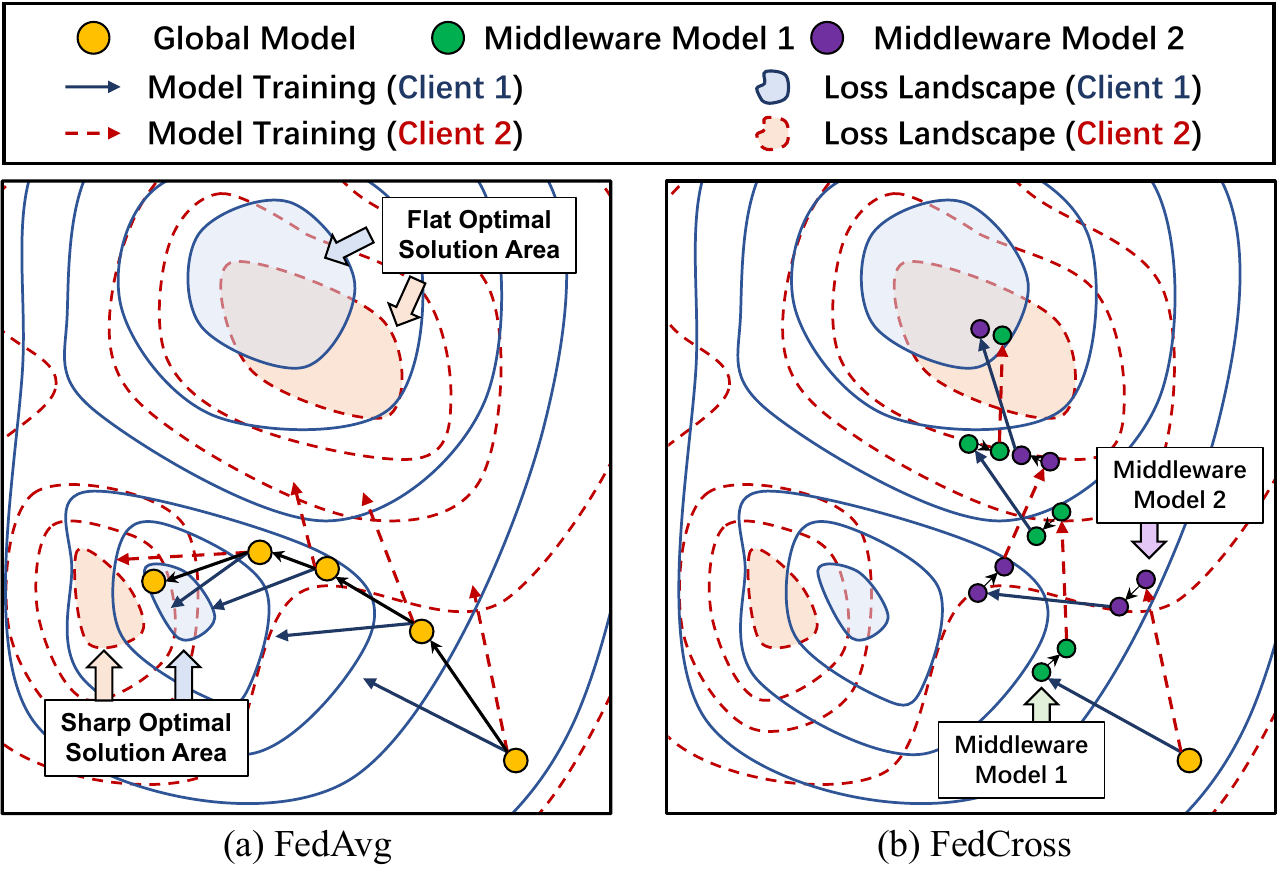}
  \vspace{ -0.15 in}
		\caption{A motivating example of FedAvg and FedCross training.}
		\label{fig:motivation} 
	\end{center}
 \vspace{ -0.15 in}
\end{figure}

According to \cite{hochreiter1997flat,hardt2016train,tsuzuku2020normalized,cha2021swad},
a well-generalized DNN training solution tends to be located in flat valleys rather than sharp ravines from the perspective of \textit{loss landscapes} \cite{hochreiter1997flat}. 
Inspired by this observation, designing an FL method to guide client model training towards a flatter valley to achieve a more generalized global model would be wise. As a motivating example, Figure~\ref{fig:motivation}(a) presents the loss landscapes 
of FedAvg involving
two clients, where blue (solid) and red (dotted) contours indicate the loss landscapes of client 1 and client 2, respectively. 
Here, we assume that 
each client has two optimal solutions (i.e., the sharp and flat optimal solutions), where the blue and red shaded areas are for client 1 and client 2, respectively. 
Note that from the perspective of loss landscapes,  a larger overlap exists between optimal solution areas if clients' data are more similar.
%
Here, we use yellow circles to denote intermediate aggregated global models
along the FL training process, where 
the black solid arrow lines form the optimization route of the global model.
We can find that  
the global model converges into the blue sharp solution area. In this case, the remaining FL  training process will inevitably get stuck in this area due to the one-to-multi style aggregation. In this case, although the obtained global model works well for client 1, it is unsuitable for client 2, although the global model is located near  (rather than in) the red-shaded area,  resulting in an inferior global model with bad generalization. 

Ideally, we can achieve a better global model for FL if local models can access the raw data of all the clients. However, this will violate the privacy-preserving requirement of FL, since both raw data and their distributions of clients are assumed to be private. Without such information, existing FedAvg-based FL methods can only tune the conflicting parameters by coarse-grained aggregations during the FL training, where the conflicting parameters of locally trained models are not properly treated. 
Apparently, {\it how to break through the limit of FedAvg to enable the fine-grained training of local models and wisely resolve the conflicting gradients to generate a well-generalized global model that performs well in all the clients with different data distributions} is becoming an urgent issue in FL design.

To address the challenge above, 
  this paper presents a novel FL framework named FedCross based on our proposed multi-model cross-aggregation-based training scheme, where we adopt  {\it middleware models} to  simultaneously  respect the local training of clients and increase
  the chance of accessing 
 different clients' data. 
Figure~\ref{fig:motivation} illustrates 
the basic idea of our FedCross approach, 
where the training process is gradually  
optimized towards the flat solution areas.
%
In this example, two middleware models are trained by the two clients (i.e., green circles for middleware model 1 and purple circles for middleware model 2). 
Note that, during FedCross training, the  
middleware models are sufficiently trained 
on different devices, indicated by interleaved
arrow lines with different colors. This way, the conflicting parameters of these two middleware models are gradually
revised by continuous local training. Eventually, their optimization directions will converge towards the intersection of flat optimal solution areas.

Unlike one-to-multi style FedAvg, FedCross conducts the local training in a multi-to-multi manner, which uses multiple middleware models to resolve the conflicts among local models on the cloud server. Rather than eliminating the conflicts immediately through FedAvg-like coarse-grained aggregation, FedCross effectively solves them by consecutive local training on different clients.
Specifically, in each training round of FedCross, the cloud server dispatches multiple homogeneous middleware models to the selected clients for local training.
After receiving all the locally trained models, FedCross applies our multi-model cross-aggregation strategy, which updates each middleware model on the cloud server by aggregating it with its collaborative model trained on some selected client.
With our multi-to-multi training scheme, each middleware model in FedCross is updated with data from different clients without privacy leaking. The conflicting weights of each middleware model can be revised by fine-grained local training rather than coarse-grained averaging aggregation. Thus, FedCross can generally achieve better classification performance than FedAvg-based FL methods. 
Due to the same set of host clients and our proposed cross-aggregation strategy that restricts the weight differences between middleware models, 
the trained middleware models will eventually become similar. 
Note that, at the end of FL training, 
FedCross only performs the federated averaging operation once on all the trained middleware models so as to form a unified ``global'' model to benefit all the clients.

This paper makes the following four major contributions:

\begin{itemize}
\item
We establish a novel multi-to-multi FL framework named FedCross, which adopts only-for-training middleware models to generate a well-generalized global model.

 
\item 
We design a multi-model cross-aggregation scheme, which supports the fine-grained training of local models to wisely resolve the conflicts among their parameters.



\item 
We  prove the convergence of FedCross and propose 
two optimization methods to accelerate the FedCross training.

\item  We conduct extensive experiments to evaluate the performance and pervasiveness of our FedCross approach.
\end{itemize}

The rest of this paper is organized as follows. Section \ref{background} introduces the preliminaries and related works on FL. Section~\ref{approach} presents the details of our proposed FedCross approach. 
Section \ref{experiment} empirically studies the performance of our FedCross approach, compared with state-of-the-art FL methods. Finally, Section \ref{conclusion} concludes the paper.

\section{Preliminaries and  Related Work}
\label{background}

\subsection{Preliminaries}

Consider learning a predictive model that maps an input space $X$ to an output space $Y$.
Assume that there 
are two entities involved in an FL system: a cloud server $S$ and $N$ distributed clients with indices of  $\{1, 2, \cdots, N\}$. 
Let each client $i$ possess a local dataset 
$D_i=\{z_{i,1}, z_{i,2}, \cdots, z_{i,n_i}\}$, where $z_{i,j}=(x_{i,j}, y_{i,j})\in X\times Y$. Under the coordination of the cloud server, all participant clients collaboratively train a global model $\hat{w}$ by sharing their local models 
trained on 
their private datasets. The goal of a standard FL optimization problem is formulated as follows:
\begin{equation*}
\footnotesize
\begin{split}
\min_{w} F(w) = \frac{1}{N}\sum_{i = 1}^{N} f_i(w), 
\ s.t.,  \ f_i(w) = \frac{1}{n_i} \sum_{j = 1}^{n_i} l (z_{i,j};w),
\end{split}
\end{equation*}
%
where $l$ and  $f_i$ denote the loss functions of an individual sample (e.g., the cross-entropy loss) and all the samples of client $i$, respectively.  $F$ represents the loss function of the global model. 
The traditional one-to-multi FL system solves this problem based on 
iterative stochastic optimization, 
where each training iteration $t$ involves four major steps:
i) model dispatching, where the cloud server selects a subset of clients and dispatches the current model $w_t$ to them; 
ii) local updating, where each selected client $i$ independently trains a local model based on $w_{t+1}^i=w_t^i-\eta \nabla F(w_t)$;  
iii) model uploading, where  each client $i$ uploads the updated local model $w_{t+1}^i$ to the cloud server; and  
iv) model aggregation, where  the server aggregates all the received models and conducts the model aggregation to obtain a new global model $w_{t+1}$ by
 FedAvg \cite{mcmahan2017communication}.


\subsection{Related Work on FL Optimization}\label{sec:relwork}

To support efficient FL in the 
 design of AIoT 
applications,  various framework- and workflow-level
optimization techniques  have been extensively
studied, including cloud-client collaboration \cite{zhang2020efficient,uddin2020mutual,zhang2021incentive,hu2023gitfl}, 
resource allocation and task scheduling \cite{luo2021cost,lim2021decentralized,cui2021client,yan2023have}, 
heterogeneity management \cite{liu2022enhancing,yang2021characterizing,li2021hermes,liu2023adapterfl,jia2023adaptivefl}, 
fault tolerance
\cite{li2021efficient,rey2022federated,li2023fedcss, li2021privacy}, and personalized service \cite{wu2021hierarchical}. 
Although these methods are promising, they can only deal with specific AIoT scenarios.
So far, to improve the classification performance of general-purpose FL methods, especially for non-IID scenarios, existing optimization methods for FL can be mainly classified into the following three categories.


The {\it global control variable-based methods}~\cite{karimireddy2020scaffold,huang2021personalized} attempt to use a global variable to guide the training direction of local training, thereby alleviating gradient divergence. 
For example, SCAFFOLD~\cite{karimireddy2020scaffold} dispatches global control variables to clients to correct the ``client-drift'' problem in the local training process.
FedProx~\cite{li2020federated} regularizes local loss functions with a proximal term to stabilize the model convergence,
where such a proximal term is the squared distance between local and global models.
The {\it client grouping-based methods}~\cite{chen2020fedcluster,fraboni2021clustered}  group clients based on the similarity of their data distributions and select clients to participate in FL training by group.
Since it is hard to directly obtain the data distributions of clients, most existing methods conduct the client grouping only based on simple information such as model similarity.
For example, 
FedCluster~\cite{chen2020fedcluster} groups the clients into multiple clusters that perform federated learning cyclically in each learning round.  
CluSamp~\cite{fraboni2021clustered} uses either 
the sample size or model similarity to group clients, which can reduce the variance of client stochastic aggregation parameters in FL.
Unlike the former two categories, the {\it Knowledge Distillation (KD)-based methods}~\cite{zhang2022fine,lin2020ensemble,ozkara2021quped}  adopt a ``teacher model'' to
guide the training of ``student models''. Specifically,
the ``student models'' use soft labels of the teacher model to perform model training, thus learning
the knowledge of the teacher model.
For example, 
FedAUX~\cite{sattler2021fedaux} performs data-dependent distillation by using an auxiliary dataset to initialize the server model.
FedGen~\cite{zhu2021data} performs data-free distillation and leverages a proxy dataset to address the heterogeneous FL problem using a built-in generator. FedDF~\cite{lin2020ensemble} uses ensemble distillation to accelerate FL by training the global model through unlabeled data on the outputs of local models.

Although various optimization methods have been proposed to 
improve  FL performance, 
due to the usage of the same global models for local training,
most of them suffer from the problem of getting stuck in sharp ravines 
during the exploration of loss landscapes. 
As an alternative, our FedCross approach adopts multiple intermediate models for local training. In this case, 
 intermediate models can quickly escape from sharp ravines based on our proposed 
cross-aggregation mechanism.
%
To the best of our knowledge, FedCross is the first attempt that uses a novel multi-to-multi training scheme based on our proposed multi-model cross-aggregation.
By using a more fine-grained FL training strategy, FedCross fully respects the convergence characteristics of clients during the training, thus achieving much better classification performance than state-of-the-art FL methods. 


\section{Our FedCross Approach}
\label{approach}
\subsection{Overview of FedCross}
The architecture of FedCross consists of a central cloud server and multiple local devices, which is the same as conventional one-to-multi FL frameworks. 
The main difference is that FedCross uses a multi-to-multi training and aggregation mechanism. Specifically, FedCross uses multiple homogeneous middleware models for local training and updates these middleware models with a cross-aggregation strategy.
FedCross still generates a global model, but this global model is only for deployment rather than model training.

\begin{figure}[h] 
  \vspace{ -0.1 in}
	\begin{center} 
		\includegraphics[width=0.45\textwidth]{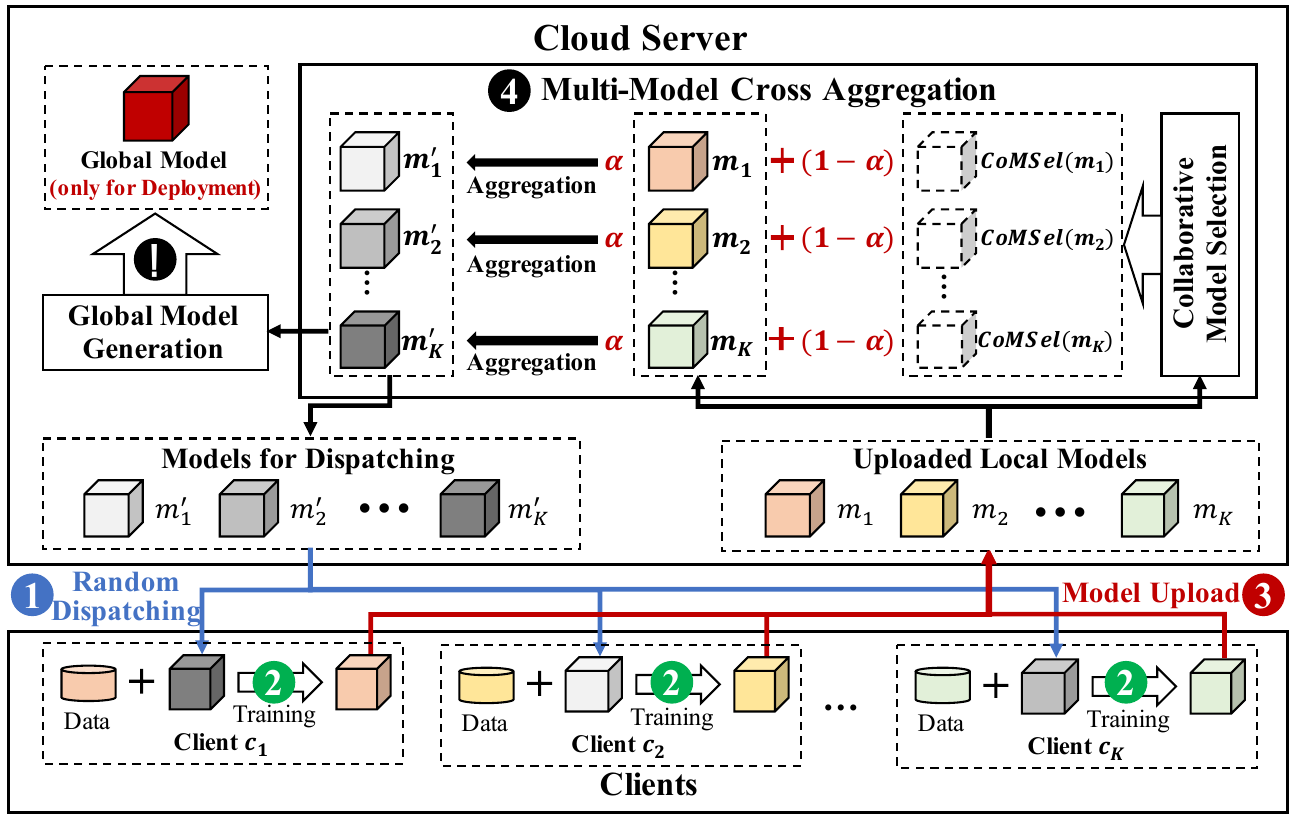}
  \vspace{ -0.1 in}
		\caption{The FedCross Framework.}
		\label{fig:framework} 
	\end{center}
 \vspace{ -0.15 in}
\end{figure}

Figure~\ref{fig:framework} presents the framework for FedCross, which shows the two processes above, i.e., model training and global model generation.
Assume that there are a total of $N$ clients. In each FL round, there are $K$ clients participating in local training, where $K\leq N$. 
The model training process trains middleware models, which consists of 4 steps:
\begin{itemize}[leftmargin=*]
    \item \textbf{Step 1 (Middleware Model Dispatching):}
    The cloud server randomly dispatches $K$ middleware models to $K$ local clients,  where each client receives one middleware model.
    
    \item \textbf{Step 2 (Middleware Model Training):} Clients train their received middleware models independently with local data.   
    \item \textbf{Step 3 (Model Uploading):} 
    All the clients upload their trained middleware models to the cloud server.
    
    \item \textbf{Step 4 (Multi-Model Cross-Aggregation):} 
    For each middleware model $m_i$ ($ 1\leq i \leq K$), FedCross chooses another middleware model $m_j$ ($j \neq i$) as the collaborative model.
    By aggregating each middleware model and its collaborative model with weights of $\alpha$ and $1-\alpha$, respectively, the cloud server generates $K$ new middleware models (i.e., $m^{\prime}_1, ..., m^{\prime}_K$) for the next-round training. 
\end{itemize}
The global model generation process aggregates multiple trained middleware models to generate a global model.
Since in FedCross the global model is only used for the model deployment, the global model generation does not need to be performed in every FL round.



\subsection{The FedCross Algorithm}
Algorithm~\ref{alg:fedcross} presents the pseudo-code of our FedCross approach.
Line~\ref{line:init} initializes a list, $W$, of $K$ dispatched models.
Lines~\ref{line:trainStart}-\ref{line:trainEnd} present the model training process.
Line~\ref{line:clientSel} randomly selects $K$ clients for each round's model training, where $L_c$ is the list of selected clients.
Line~\ref{line:clientShuffle} shuffles the order of the selected models, with which each dispatched model is given an equal chance to be trained by the client. Note that, without shuffling, each middleware model will be dispatched to the clients encountered in the previous training rounds with a high probability.
Lines~\ref{line:localTrainStart}-\ref{line:localTrainEnd} dispatch models to the corresponding clients and conduct local training process. In Line~\ref{line:clientUpdate}, each client trains the received model using local data and uploads the retrained local model to the cloud server. In Line~\ref{line:modelUpdate}, the cloud server updates the model list $W$ using the received trained model.
In Line~\ref{line:comodelsel}, the function {\it CoModelSel} selects a collaborative model for each uploaded model.
In Line~\ref{line:crossAggr}, the function {\it CrossAggr} aggregates each uploaded model with its collaborative model to generate $K$ models.
Line~\ref{line:wupdate} updates the dispatched model list $W$ using these generated models.
In Line~\ref{line:modelAggr}, the function {\it GlobalModelGen} generates a global model for the deployment by aggregating all the models in $W$.
Since the global model does not participate in the model training, the global model generation can be performed asynchronously at any time.
The following will detail the key parts of FedCross and analyze its convergence.

\setlength{\textfloatsep}{5pt}
\begin{algorithm}[ht]
\caption{The FedCross Algorithm}
\label{alg:fedcross}
\KwIn{
    {\bf i)} $round$, \# of training rounds;
    {\bf ii)} $C$, the set of clients;
    {\bf iii)} $K$, \# of clients  participating in each FL round.
 }
 \KwOut{
   $w_{g}$, the global model.
 }
\textbf{FedCross}($round$,$C$,$K$) \Begin{
$W\leftarrow [w^1_0, w^2_0,...,w^{K}_0]\ \ \ $ // initialize the model list\;\label{line:init}
\For{$r$ = 0, ..., $round-1$}{\label{line:trainStart}
    $L_c\leftarrow$ Random select $K$ clients from $C$\;\label{line:clientSel}
    $L_c\leftarrow Shuffle(L_c)$\;\label{line:clientShuffle}
    /*parallel for block*/\\
    \For{$i$ = 1, ..., $K$}{ \label{line:localTrainStart}
        $v_{r+1}^i\leftarrow ${\it LocalTraining}$(w^{i}_r,L_c[i])$\;\label{line:clientUpdate}
        $W[i]\leftarrow v_{r+1}^i$\;\label{line:modelUpdate}
    } \label{line:localTrainEnd}
    \For{$i$ = 1, ..., $K$}  {
        $v_{co}^i\leftarrow ${\it CoModelSel}$(v^i_{r+1},W)$\label{line:comodelsel}\;
        $w_{r+1}^i\leftarrow ${\it CrossAggr}$(v^i_{r+1},v_{co}^i)$\;\label{line:crossAggr}
    }
    $W\leftarrow [w_{r+1}^1,w_{r+1}^2,..., w_{r+1}^K]$\;\label{line:wupdate}
}\label{line:trainEnd}
$w_{g}\leftarrow ${\it GlobalModelGen}$(W)$\label{line:modelAggr}\;
{\bf return} $w_{g}$\;
}
\end{algorithm}

\subsubsection{Collaborative Model Selection ({\it CoModelSel})}
To facilitate knowledge exchange between models, FedCross selects a collaborative model for each in the uploaded model list for cross-aggregation.
According to model characteristics, we design   three following
model selection criteria to accommodate different purposes: i) \textit{adequacy-and-diversity of participation}, ii) \textit{minimizing gradient divergence}, and iii) \textit{maximizing the knowledge acquisition}. 

Since each middleware model is trained on a client, the knowledge acquired by each model is different.
To fully exploit the information in the uploaded models, the \textit{adequacy-and-diversity criterion} encourages each model to update other models as much as possible. This way, each middleware model can acquire diverse knowledge.
Based on this criterion, we ordinally select a collaborative model from the middleware model list for the target model.

Since middleware models trained on different clients inevitably have differences, the \textit{gradient divergence minimization criterion} encourages each model to find a similar collaborative model for the cross-aggregation to minimize gradient divergence in each cross-aggregation.
Based on this criterion, we present \textit{the highest similarity} strategy, which selects the most similar model to the target model.

The \textit{knowledge maximization criterion} encourages each model to obtain more knowledge at each training round.
Since models with high similarity have similar knowledge, contrary to the gradient divergence minimization criteria, the knowledge maximization criteria prefer to select a model with low similarity to the target model.
Based on this criterion, we present \textit{the lowest similarity} strategy, which selects the least similar model for the target model.
%
%
%
%
The details of the three model selection strategies (i.e., in-order, highest similarity, lowest similarity) are as follows:

\textbf{In-order strategy:} 
For the $i^{th}$ model, the cloud server selects the $((i+(r\%(K-1)+1))\%K)^{th}$ model as the collaborative model in the $r^{th}$ training round. The  in-order strategy is as follows:
\begin{equation}
\footnotesize
\begin{split}
CoModelSel(v^i_{r},W) =  W[(i+(r\%(K-1)+1))\%K],
\end{split}
\nonumber 
\end{equation}
where $W$ is the list of uploaded local model parameters, and $K$ is the number of uploaded models.
With this strategy, all the upload models are chosen as collaborative models in each round. Note that, in every $(K-1)$  rounds
of training, each middleware model collaborates with all the other $(K-1)$ models once.

\textbf{The highest similarity strategy:}
By calculating the model similarity between the uploaded models, each middleware model aggregates the model with the highest similarity as follows:
\begin{equation}
\footnotesize
\begin{split}
CoModelSel(v^i_{r},W) = \underset{v\in W\setminus\{v^i_{r}\}}{\arg\max}\ Similarity(v^i_{r},v),
\end{split}
\nonumber 
\end{equation}
where $W$ is a list of uploaded local model parameters and 
{\it Similarity($\cdot$)} is a function to calculate the model similarity.
Note that a higher {\it Similarity($\cdot$)} value means a higher similarity between the two models.

\textbf{The lowest similarity strategy:}
According to the definition of the highest similarity strategy, the lowest similarity strategy encourages each model to select the model with the least similarity as the collaborative model:
\begin{equation}
\footnotesize
\begin{split}
CoModelSel(v^i_{r},W) = \underset{v\in W\setminus\{v^i_{r}\}}{\arg\min}\ Similarity(v^i_{r},v).
\end{split}
\nonumber 
\end{equation}

In this paper, since the classic  cosine similarity
 can accurately reflect the 
 angles of gradients, we adopt it as the measure as follows:
\begin{equation}
\footnotesize
\begin{split}
Similarity(X,Y)=\frac{\sum^n_{i=1}{X_i\times Y_i}}{\sqrt{\sum^n_{i=1}{X_i^2}}+\sqrt{\sum^n_{i=1}{Y_i^2}}},
\end{split}
\nonumber 
\end{equation}
where $X$ and $Y$ are two models, $n$ indicates the number of parameters, and $X_i$ indicates the $i^{th}$ parameter in $X$. We would like to leave interesting topics of using other measures (e.g., Euclidean Distance) as our future work.

Compared with both in-order and the lowest similarity strategies, there are obvious flaws in the highest similarity strategy.
Since the goal of FedCross is still to train a high-performance global model, the collaborative model selection strategy should guide all middleware models to be optimized in a similar direction. Although the {\it the highest similarity} strategy makes the lowest gradient divergence in each cross-aggregation, from a global perspective, such strategy makes models with high similarity increasingly similar, and it is more and more difficult for dissimilar models to share knowledge.
At the end of FL training, middleware models are clustered into several groups, and the optimization directions of such groups are different.
Finally, in the deployment phase, more serious gradient conflicts than ever will occur in the aggregation of the global model.

%




\subsubsection{Cross-Aggregation ({\it CrossAggr})} 
The cross-aggregation is a novel multi-to-multi aggregation method, which fuses each upload model with its collaborative model with the weight  $\alpha$.
Suppose that $v^i_{r}$ is an uploaded model and $v^i_{co}$ is its collaborative model. The cross-aggregation process is as follows:
\begin{equation}
\footnotesize
\begin{split}
CrossAggr(v^i_{r},v^i_{co}) = \alpha\times v^i_{r} + (1-\alpha)\times v^i_{co},
\end{split}
\nonumber 
\end{equation}
where $\alpha\in [0.5,1.0)$ is a hyperparameter used to determine the weight of the aggregation. The adjustment of $\alpha$ is important and difficult. If $\alpha$ is small, the gradient conflict will become serious. If $\alpha$ is large, it is difficult for the model to learn the knowledge of the collaborative model. Thus, we conduct an ablation study to confirm the reasonable value space of $\alpha$ by 
evaluating the performance of FedCross with different $\alpha$ values in Section~\ref{sec:exp_criteria}.


\subsubsection{Global Model Generation}
The global model generation phase is the same as the traditional FL methods. In FedCross, the global model does not participate in model training and is only used for model deployment. Thus, the global model can be performed asynchronously with model training. The global model is obtained by the following formula:
\begin{equation}
\footnotesize
\begin{split}
w_{g} = \frac{1}{K}\sum_{i = 1}^{K} w_{r}^{i}
\nonumber 
\end{split}
\end{equation}
where $w_{r}^i$ is the parameters of the $i^{th}$  model in the dispatched model list, and $r$ is the number of the current training round.

\subsection{Convergence Analysis}

Inspired by the proof of the convergence of traditional one-to-multi FL approach~\cite{convergence,karimireddy2020scaffold}, we prove the convergence of FedCross as follows.

\subsubsection{Notations}
Assume that all clients adopt Stochastic Gradient Descent (SGD) as the optimizer.
Let $t$ be the number of rounds of the current SGD iteration on clients, 
and 
$w^i_t$ be the parameters of the $i^{th}$ middleware model. 
After exactly one  SGD iteration, we can get the parameters of some local model,
i.e., $v^i_{t+1}$, by using the following  model update formula:
%
\begin{equation}
\footnotesize
\begin{split}
v_{t+1}^i = w_t^i - \eta_t \nabla f_i (w_t^i, \xi_t^i), 
\end{split}
\nonumber 
\end{equation}
Assuming that each local model is uploaded to the cloud server in every
$E$  iterations and  $i^\prime = (i + (t\%E)\%(N-1)+1)\%N$, we have
\begin{equation}
\footnotesize
\begin{split}
w_{t+1}^i=\left\{
\begin{array}{rl}
v_{t+1}^i, &if (t + 1) \% E \neq 0 \\
\alpha v_{t+1}^i + (1-\alpha)v_{t+1}^{i^\prime}, &if  (t + 1) \% E = 0\\
\end{array},
\right.  
\end{split}
\nonumber 
\end{equation}
%
Since FedCross generates 
a global model 
by aggregating all the middleware models, we use
two variables $\overline{v}_t$ and $\overline{w}_t$ to represent
the aggregated model of all  middleware models:
\begin{equation}
\footnotesize
\begin{split}
    \overline{v}_t = \frac{1}{N} \sum_{i=1}^{N} v_t^i, \ 
    \overline{w}_t = \frac{1}{N} \sum_{i=1}^{N} w_t^i.
\end{split}
\nonumber 
\end{equation}
We define $g^i_t$ to denote the gradients of the model in the $i^{th}$ client after training with a data batch $\xi_t^i$:
\begin{equation}
\begin{split}
    g^i_t =\nabla f_i (w_t^i; \xi_t^i).
\end{split}
\nonumber 
\end{equation}


\subsubsection{Proofs of Key Lemmas}
We analyze the convergence of FedCross 
based on three assumptions for the loss function of each client (i.e., $f_1, f_2, ...,$ or $f_N$), including $L$-smooth assumption (Assumption \ref{asm1}), $\mu$-convex assumption (Assumption \ref{asm2}), and variance/mean bound assumption for stochastic gradients (Assumption \ref{asm3}), which have been used in prior works \cite{ zhang2012communication, stich2018local, convergence}.

\begin{assumption}\label{asm1}
$f_i$ is $L$-smooth satisfying $|| \nabla f_i(w) - \nabla f_i(w^\prime) || \leq L ||w - w^\prime||$, where $i \in\{ 1, 2, \cdots, N\}$.
\end{assumption}
\begin{assumption}\label{asm2}
$f_i$ is $\mu$-convex satisfying $|| \nabla f_i(w) - \nabla f_i(w^\prime) || \geq \mu||w - w^\prime||$, where $i \in\{ 1, 2, \cdots, N\}$ and $\mu \geq 0$.
\end{assumption}
\begin{assumption}\label{asm3}
The variance of stochastic gradients is upper bounded by  $\sigma^2$ and the expectation of squared norm of stochastic gradients is upper bounded by  $G^2$, i.e.,  $\mathbb{E}||\nabla f_i (w;\xi) - \nabla f_i (w) ||^2 \leq \sigma^2$, $\mathbb{E}||\nabla f_i (w;\xi) ||^2 \leq G^2$, where $\xi$ is a  data batch of the $i^{th}$ client in the $t^{th}$ FL round.
\end{assumption}

Assume that in FedCross
all the $N$ clients are participating in every FL training round, and we employ
the in-order selection strategy.
Let $\{v^1_r,v^2_r,..,v^N_r\}$ 
be the set of uploaded local model parameters in the $(r-1)^{th}$ round, $\{w^1_r,w^2_r,..,w^N_r\}$ be the set of cross-aggregated model parameters, and $i^\prime = (i + r\%(N-1)+1)\%N$ be the index of collaborative model of the $i^{th}$ middleware model.
Based on the implementation of our in-order strategy, we have
\begin{equation}\label{eq:mr1}
\footnotesize
\begin{split}
w_{r}^i= \alpha v_{r}^i + (1-\alpha)v_r^{i^\prime}.
\end{split}
\end{equation}
Since in the in-order strategy each uploaded model is selected as a collaborative model for cross-aggregation, we have
\begin{equation}\label{eq:mr2}
\footnotesize
\begin{split}
\sum_{i=1}^N w_{r}^i= \sum_{i=1}^N(\alpha v_{r}^i + (1-\alpha)v_r^{i^\prime}) = \sum_{i=1}^N v_{r}^i
\end{split}
\end{equation}
According to Equations~\ref{eq:mr1}-\ref{eq:mr2}, we have Lemma~\ref{eq:lemma1} as follows.
\begin{lemma}\label{eq:lemma1} Let $w_{r}^i= \alpha v_{r}^i + (1-\alpha)v_r^{i^\prime}$, $\alpha\in [0,1]$, and $\overline{w}_r = \sum_{i=1}^N w_{r}^i$. We have
\begin{equation}
\footnotesize
\begin{split}
||\overline{w}_r - w^\star||^2\leq\frac{1}{N}\sum_{i=1}^N||w_{r}^i-w^\star||^2\leq\frac{1}{N}\sum_{i=1}^N||v_{r}^i-w^\star||^2,
\nonumber
\end{split}
\end{equation}
where $w^\star$ is the optimal parameters for the global loss function $F(\cdot)$. In other words, $\forall w, F^\star\leq F(w)$, where $F^\star$ denotes $F(w^\star)$. 
\end{lemma}

\begin{proof} 
We can derive the following inequality:
\begin{equation*}
\footnotesize
\begin{split}
\sum_{i=1}^N||w_{r}^i-w^\star||^2 & = \sum_{i=1}^N||\alpha v_{r}^i + (1-\alpha)v_r^{i^\prime}-w^\star||^2\\
& = \sum_{i=1}^N(||v_r^i - w^\star||^2 - \alpha(1-\alpha)||v_r^i - v_r^{i^\prime}||^2)\\
&\leq \sum_{i=1}^N||v_r^i - w^\star||^2.
\end{split}
\end{equation*}

Since $\overline{v}_r= \overline{w}_r = \frac{1}{N}\sum_{i=1}^N w_r^i$ holds, by using the AM–GM inequality, we can obtain:
\begin{small}
\begin{equation*}
||\overline{v}_r-w^\star||^2 \leq \frac{1}{N}\sum_{i=1}^N ||w_r^i-w^\star||^2.
\end{equation*}.
\end{small}
\end{proof}

To facilitate the convergence analysis of FedCross, we present Lemmas \ref{eq:lemma2}-\ref{eq:lemma3}.
\begin{lemma} \label{eq:lemma2} (Results of one  step SGD). If $\eta_t\leq \frac{1}{4L}$ holds, we have:
\begin{small}
\begin{equation*}
\footnotesize
\begin{split}
    \mathbb{E}||\overline{v}_{t+1} - w^\star||^2 \leq &\frac{1}{N}\sum_{i=1}^N(1-\mu\eta_t)||w^i_t -w^\star||^2 \\
    & + \frac{1}{N}\sum_{i=1}^N||w^i_t - w^i_{t_0}||^2 + 10\eta_t^2 L\Gamma.
\end{split}
. \nonumber
\end{equation*}
\end{small}
\end{lemma}

\begin{proof} 
By using the AM–GM inequality, it holds that:
\begin{equation*}
\footnotesize
\begin{split}
    ||\overline{v}_{t+1} - w^\star||^2 &
    \leq\frac{1}{N}\sum_{i=1}^N||v^i_{t+1} - w^\star||^2\\
     & = \frac{1}{N}\sum_{i=1}^N(||v^i_t -w^\star||^2 -2\eta_t \langle v^i_t-w^\star, g^i_t \rangle
     + \eta_t^2||g^i_t||^2).
\end{split}
\end{equation*}
Let $P_1 =  -2\eta_t\langle w^i_t-w^\star, g^i_t \rangle$ and $P_2=\eta_t^2\sum_{i=1}^N||g^i_t||^2$.
By using $\mu$-convex (Assumption \ref{asm2}), we have:
\begin{equation}\label{eq_b1}
\footnotesize
\begin{split}
P_1 \leq -2\eta_{t}(f_i (v^i_t)-f_i (w^\star))-\mu \eta_t ||w^i_t-w^\star||^2.
\end{split}
\end{equation}
By using $L$-smooth (Assumption \ref{asm1}), we obtain:
\begin{equation}\label{eq_b2}
\footnotesize
\begin{split}
P_2 \leq 2\eta_t^2 L (f_i(w_t^i)-f_i^\star).
\end{split}
\end{equation}
When $(t+1)\%E\neq 0$ and $v_t^i=w_t^i$ hold, according to Equations \ref{eq_b1}-\ref{eq_b2}, we have:
\begin{equation*}
\footnotesize
\begin{split}
    ||\overline{v}_{t+1} - w^\star||^2 
    \leq & \frac{1}{N}\sum_{i=1}^N [(1-\mu\eta_t)||v^i_t -w^\star||^2  -2\eta_{t}(f_i (w^i_t)-f_i (w^\star)) \\&+2\eta_t^2 L(f_i(w_t^i)-f_i^\star)].
\end{split}
\end{equation*}
    Let $P_3 = \frac{1}{N}\sum_{i=1}^N[-2\eta_{t}(f_i (w^i_t)-f_i (w^\star)) + 2\eta_t^2 L (f_i(w_t^i)-f_i^\star)]$. It holds that:
\begin{equation*}
\footnotesize
\begin{split}
    P_3 
    & = -\frac{2\eta_t(1-\eta_t L)}{N}\sum_{i=1}^N(f_i(w^i_t)-F^\star) + \frac{2\eta_t^2 L}{N}\sum_{i=1}^N(F^\star-f^\star_i).
\end{split}
\end{equation*}
Let $\Gamma=F^\star-\frac{1}{N}\sum_{i=1}^N f^\star_i$ and $\phi=2\eta_t(1-L\eta_t)$. We have:
\begin{equation*}
\footnotesize
\begin{split}
    P_3 = -\frac{\phi}{N}\sum_{i=1}^N(f_i(w^i_t)-F^\star) + 2\eta_t^2 L\Gamma.
\end{split}
\end{equation*}
Let $P_4=-\frac{1}{N}\sum_{i=1}^N(f_i(w^i_t)-F^\star)$, $t_0 \% E = 0$ and $t-t_0\leq E$. It holds that:
\begin{equation*}
\footnotesize
\begin{split}
    P_4 = -\frac{1}{N}\sum_{i=1}^N(f_i(w^i_t) - f_i(w^i_{t_0}) + f_i(w^i_{t_0}) -F^\star)
\end{split}.
\end{equation*}
Based on the Cauchy–Schwarz inequality, we can derive that:
\begin{equation}\label{eq_D}
\footnotesize
\begin{split}
    P_4 \leq & \frac{1}{2N}\sum_{i=1}^N(\eta_t ||\nabla f_i(w^i_{t_0})||^2 + \frac{1}{\eta_t}||w^i_t - w^i_{t_0}||^2)\\&  -\frac{1}{N}\sum_{i=1}^N(f_i(w^i_{t_0}) -F^\star)\\
     \leq& \frac{1}{2N}\sum_{i=1}^N \left[2\eta_t L(f_i(w^i_{t_0})-f_i^\star) + \frac{1}{\eta_t}||w^i_t - w^i_{t_0}||^2\right]\\& - \frac{1}{N}\sum_{i=1}^N(f_i(w^i_{t_0}) -F^\star).
\end{split}
\end{equation}

Note that, since $\eta\leq \frac{1}{4L}$, $\eta_t \leq \phi\leq 2\eta_t$ and $\eta_t L \leq \frac{1}{4}$, 
according to Equation \ref{eq_D}, we have:
\begin{equation*}
\footnotesize
\begin{split}
    P_3 &\leq\frac{\phi}{2N}\sum_{i=1}^N \left[2\eta_t L(f_i(w^i_{t_0})-f_i^\star) + \frac{1}{\eta_t}||w^i_t - w^i_{t_0}||^2\right]\\
    &- \frac{\phi}{N}\sum_{i=1}^N(f_i(w^i_{t_0}) -F^\star) + \eta_t^2 L\Gamma\\
    & \leq \frac{\phi}{2\eta_t N}\sum_{i=1}^N||w^i_t - w^i_{t_0}||^2 + (\phi \eta_t L + 2\eta_t^2 L)\Gamma + \frac{\phi}{N}\sum_{i=1}^N(F^\star - f_i^\star)\\
    &\leq  \frac{1}{N}\sum_{i=1}^N||w^i_t - w^i_{t_0}||^2 + 10\eta_t^2 L \Gamma.
\end{split}
\end{equation*}
\end{proof}

\begin{lemma} \label{eq:lemma3}
In FedCross, the cross-aggregation occurs every $E$ iteration. For arbitrary $t$, there always exists $t_0 \leq t$ while $t_0$ is the nearest cross-aggregation
 to $t$.  As a result, $t - t_0 \leq E-1$ holds. Given the constraint on learning rate from \cite{convergence}, we know that $\eta_t \leq \eta_{t_0} \leq 2 \eta_t$. It follows that:
\begin{equation*}
\footnotesize
\begin{split}
 \frac{1}{N} \sum_{i=1}^{N} ||w_t^i - {w}^i_{t_0}||^2 \leq 4\eta_t^2 (E - 1)^2 G^2.
\end{split}
\nonumber
\end{equation*}
\end{lemma}

\begin{proof} 
Let $t_0\% E = 0$ and $t-t_0\leq E$. We have:
\begin{equation*}
\footnotesize
\begin{split}
\frac{1}{N} \sum_{i=1}^{N}||w_t^i - {w}^i_{t_0}||^2 = &\frac{1}{N} \sum_{i=1}^{N} \left|\left|\sum_{t=t_0}^{t_0 + E - 1} \eta_t \nabla f_{a_1}(w_t^{a_1};\xi_t^{a_1})\right|\right|^2   \\
& \leq  (E - 1) \sum_{t = t_0}^{t_0 + E - 1} \eta_t^2 G^2  \\
&\leq 4\eta_t^2 (E - 1)^2 G^2.
\end{split}
\nonumber
\end{equation*}
\end{proof}

Based on  Lemmas~\ref{eq:lemma1}-\ref{eq:lemma3}, we prove Theorem \ref{thm1} as follows.

\newtheorem{thm}{\bf Theorem}
\begin{thm}\label{thm1}
Let $E$ be the number of SGD iterations conducted within one FL round, and the whole training consists of $r$ FL rounds. Let $t=r\times  E$ be the total number of SGD iterations conducted so far, and $\eta_t=\frac{2} {\mu (t + \lambda)}$ be the learning rate. We have: 
\begin{equation}
\label{eq:thm1}
\footnotesize
\begin{split}
\mathbb{E}[F(\overline{w}_t)] -F^\star \leq \frac{L}{2\mu(t+\lambda)}\left[\frac{4B}{\mu} + \frac{\mu(\lambda+1)}{2}\Delta_1\right]
\end{split},
\end{equation}
where
\begin{small}
$
    B = 10 L \Gamma + 4(E - 1)^2 G^2.
     \nonumber
$
\end{small}
\end{thm}

\begin{proof} Let $\Delta_t = ||\overline{w}_t - w^\star||^2$ and $\Delta^{glb}_t = \frac{1}{N} \sum_{i=1}^{N}||w_t^i - {w}^\star||^2$.
According to Lemma \ref{eq:lemma1},  \ref{eq:lemma2}, and  \ref{eq:lemma3}, we have:
\begin{equation*}
\small
\begin{split}
    \Delta_{t+1} \leq \Delta^{glb}_{t+1} \leq (1-\mu\eta_t)\Delta^{glb}_{t} + \eta_t^2 B
\end{split}.\nonumber
\end{equation*}
When the step size becomes smaller, we have
 $\eta_t = \frac{\beta}{t + \lambda}$ for
 some $\beta > \frac{1}{\mu}$, $\lambda > 0$ such that $\eta_t \leq min\left\{\frac{1}{\mu},\frac{1}{4L}\right\}=\frac{1}{4L}$ and $\eta_t \leq 2\eta_{t+E}$.

Let $\theta = max\left\{\frac{\beta^2 B}{\mu\beta-1},(\lambda+1)\Delta_1\right\}$. We firstly prove  $\Delta_t\leq \frac{\theta}{t + \lambda}$ by induction.
When $t=1$,
\begin{equation}\label{eq:math_1}
\footnotesize
    \Delta_1 = \Delta^{glb}_1 = \frac{\lambda + 1}{\lambda + 1}\Delta_1 \leq \frac{\theta}{\lambda + 1}.
\end{equation}
Assuming
that $\Delta_t\leq\Delta^{glb}_t\leq \frac{\theta}{\lambda + 1}$, we have:
\begin{equation}\label{eq:math_t+1}
\footnotesize
\begin{split}
    \Delta_{t+1} & \leq \Delta^{glb}_{t+1}\\
    & \leq (1-\mu\eta_t)\Delta^{glb}_{t} + \eta_t^2 B\\
    & \leq \frac{t+\lambda - 1}{(t+\lambda)^2}\theta + \left[\frac{\beta^2 B}{(t+\lambda)^2} - \frac{\mu\beta -1}{(t+\lambda)^2}\theta\right]\\
    & \leq \frac{\theta}{t+ 1 + \lambda}.
\end{split}
\end{equation}
According to Equations \ref{eq:math_1}-\ref{eq:math_t+1},
we have:
\begin{equation}\label{eq:math_proof}
\footnotesize
\begin{split}
{\small
     \Delta_t\leq \frac{\theta}{t + \lambda}}.
     \end{split}
\end{equation}
From Assumption \ref{asm1} and Equation~\ref{eq:math_proof}, we obtain:
\begin{equation}\label{t_1}
\footnotesize
\begin{split}
\mathbb{E}[f(\overline{w}_t)] -F^\star \leq \frac{L}{2}\Delta_{t}\leq \frac{\theta L}{2(t + \lambda)}.
\end{split}
\end{equation}
If we set $\beta =\frac{2}{\mu}$ and $\lambda=max\{\frac{10L}{\mu},E\}-1$, we have $\eta_t=\frac{2}{\mu(t + \lambda)}$ and $\eta_t \leq 2\eta_{t+E}$ for $t\ge 1$. Then, it holds that:
\begin{equation}\label{t_2}
\footnotesize
\begin{split}
\theta & =max\left\{\frac{\beta^2 B}{\mu\beta-1},(\lambda+1)\Delta_1\right\}\\
     & \leq \frac{\beta^2 B}{\mu\beta-1} +(\lambda+1)\Delta_1 \\
     & \leq \frac{4B}{\mu^2} + (\lambda+1)\Delta_1.
\end{split}
\end{equation}
Based on Equations \ref{t_1}-\ref{t_2}, we have:
\begin{equation*}
\footnotesize
\begin{split}
\mathbb{E}[F(\overline{w}_t)] -F^\star & \leq \frac{L}{2(t+\lambda)}\left[\frac{4B}{\mu^2} + (\lambda+1)\Delta_1\right]\\
& = \frac{L}{2\mu(t+\lambda)}\left[\frac{4B}{\mu} + \frac{\mu(\lambda+1)}{2}\Delta_1\right].
\end{split}
\end{equation*}
\end{proof}

Theorem \ref{thm1} indicates that the difference between the current loss $F(\overline{w}_t)$ and the optimal loss $F^\star$ is inversely related to $t$. From Theorem \ref{thm1}, we observe that
as the value of $t$ increases, the right side of Equation~\ref{eq:thm1} in Theorem~\ref{thm1} will approach 0, indicating that FedCross will eventually converge.
In addition, we can also find that the convergence rate of FedCross is similar to that of FedAvg, which has been analyzed in \cite{convergence}. 


\subsection{Training Acceleration Methods
for FedCross}
Although the vanilla FedCross (i.e., FedCross without any training acceleration) can achieve the best accuracy performance compared with traditional aggregation methods (see Section \ref{sec:exp_acc}), due to our proposed fine-grained training strategy,
it still suffers from the slow convergence during FL training.
Especially in each FL training round at the early stage of  training,
due to  significant
knowledge differences among clients, the  knowledge  learned by each middleware model is limited, resulting in the
low performance of aggregated global models.
However, 
as the number of training rounds increases, each middleware model gradually
becomes well-trained with fully exchanged knowledge, leading to
a notable increase in the similarity among middleware models. Meanwhile, 
the classification performance of the global model improves significantly as well.
Note that for the cross-aggregation, the value of $\alpha$ determines how much new knowledge a model can learn from its collaborative model. Specifically, a larger $\alpha$ indicates less knowledge can be learned from its collaborative model, leading to slow convergence.

Since the fusion weight (i.e., $\alpha$) of a middleware model is much higher than that of its collaborative model in each cross-aggregation process, 
FedCross needs a large number of 
training rounds to unify all the middleware models.
To accelerate the convergence of FedCross, 
we propose two optimization methods (i.e.,
{\it propeller models} and {\it dynamic} $\alpha$) by dividing its training procedure into two stages, where the  first stage allows middleware models to learn from each other in a coarse-grained manner, while  the second stage adopts a fine-grained heuristic
to fine-tune the middleware models.
%
This way, we can balance the convergence rate and accuracy performance for a better training procedure.
The following details the 
 two training acceleration methods:

\begin{itemize}[leftmargin=*]
\item \textbf{Propeller models-based acceleration:} To fully exploit the information of uploaded middleware models, we use propeller models that are selected by the in-order selection strategy from the middleware model list.
For each middleware model, we use multiple propeller models rather than
one collaborative model to provide more knowledge that can be learned by middleware models, thus significantly accelerating the training procedure.


\item \textbf{Dynamic}  $\bm{\alpha}$\textbf{-based acceleration}: 
To accelerate the overall training convergence,
 we encourage middleware models to learn more knowledge from their collaborative models in earlier FL training rounds. 
Along with the process of  FL training, since
each middleware model can learn more knowledge with a smaller
value of $\alpha$, we 
 gradually increase
the value of $\alpha$ from 0.5 to a specific threshold (e.g., $\alpha=0.99$  used in our experiments).


\end{itemize}

\section{Experimental Results}
\label{experiment}

To evaluate the performance of FedCross, we conducted extensive experiments 
on well-known datasets and underlying DNN models. 
The subsequent subsections aim to answer the following four research questions (RQs).

\textbf{RQ1: (Validation of Motivation)}: 
Compared with FedAvg-based methods, can FedCross converge into a flatter valley?

\textbf{RQ2: (Superiority of FedCross)}: 
What are  FedCross merits  compared with 
state-of-the-art FedAvg-based methods?

\textbf{RQ3: (Compatibility of FedCross)}:
What is the performance of FedCross with different settings (e.g.,  
client data distributions, DNN architectures, datasets)?


\textbf{RQ4: (Benefits  of FedCross Components)}: 
Can our proposed techniques improve classification performance?



\subsection{Experimental Settings}
We implemented FedCross on top of vanilla
FedAVg by modifying its one-to-multi training scheme.
Similar to the work in \cite{mcmahan2017communication}, 
in the experiments,  we assumed that 
only 10\% of clients are selected to participate in the training.
To ensure  comparison fairness, for all the involved
FL methods, we set the local training batch size to 50 and performed five epochs for each local training round. 
For each client, we used SGD as the optimizer with a learning rate 
of 0.01 and a  momentum of 0.5. 
For FedCross, we set $\alpha=0.99$ and adopted the lowest similarity criterion to select collaborative models. 
We did not use other optimization methods (e.g., data augmentation) in all the following experiments.
All the experimental results were obtained from an Ubuntu workstation with  Intel i9 CPU, 32GB memory, and NVIDIA RTX 3080 GPU.

\begin{figure}[h]
	\vspace{-0.1in}
	\centering
	\subfigure[$\beta=0.1$]{
		\centering
  \hspace{-0.1in}
		\includegraphics[width=0.16\textwidth]{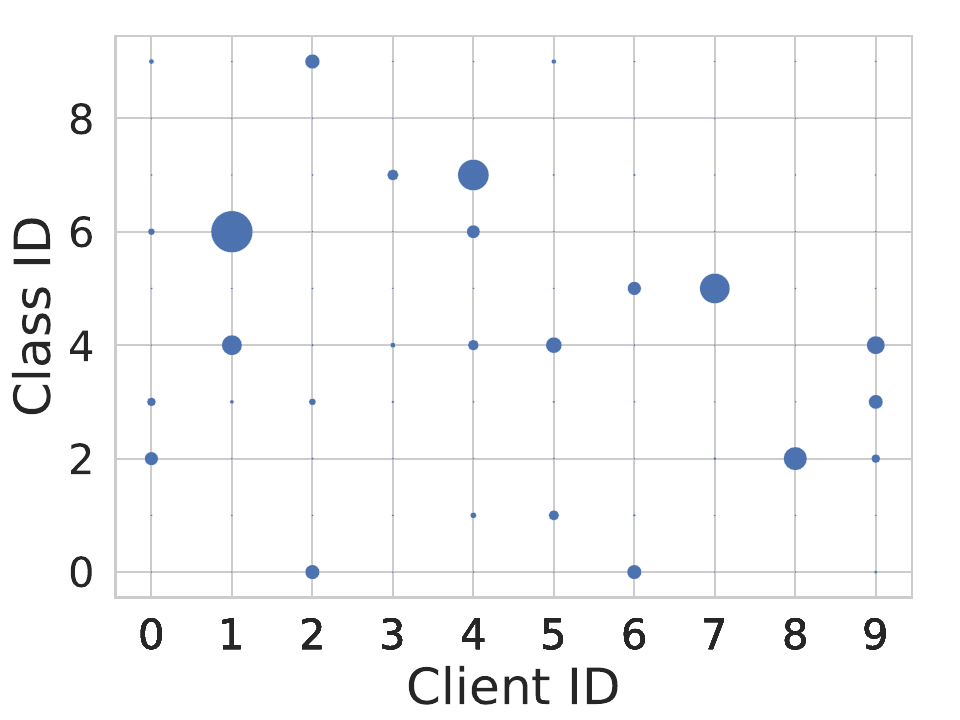}
		\label{fig:dis_d0.1}
	}\hspace{-0.15in}
	\subfigure[$\beta=0.5$]{
		\centering
		\includegraphics[width=0.16\textwidth]{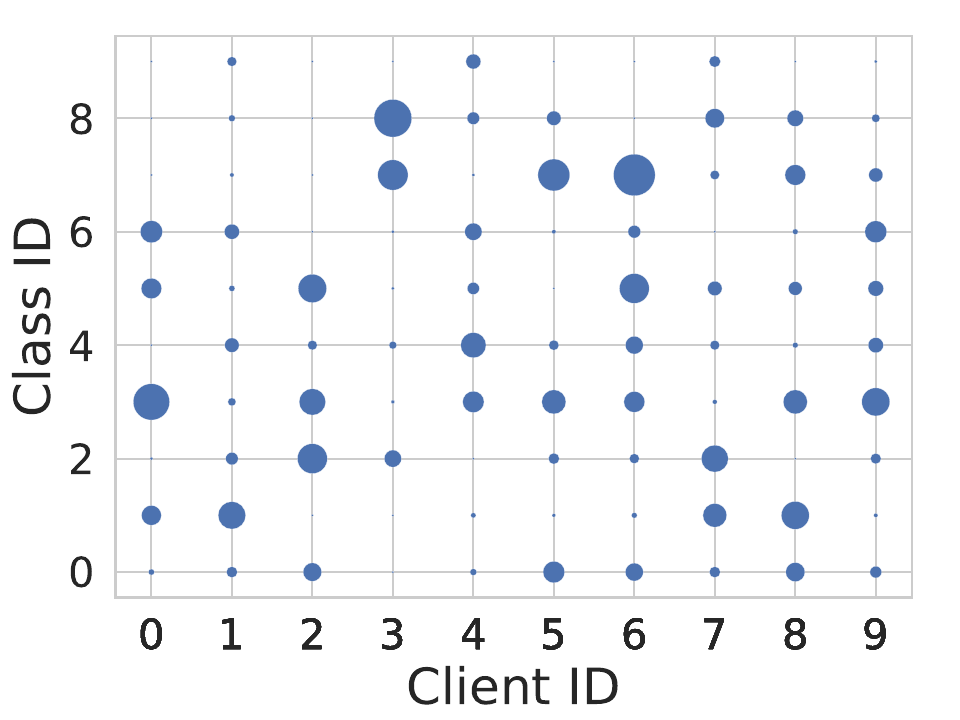}
  \label{fig:dis_d0.5}
        }\hspace{-0.15in}
        \subfigure[ $\beta=1.0$]{
		\centering
		\includegraphics[width=0.16\textwidth]{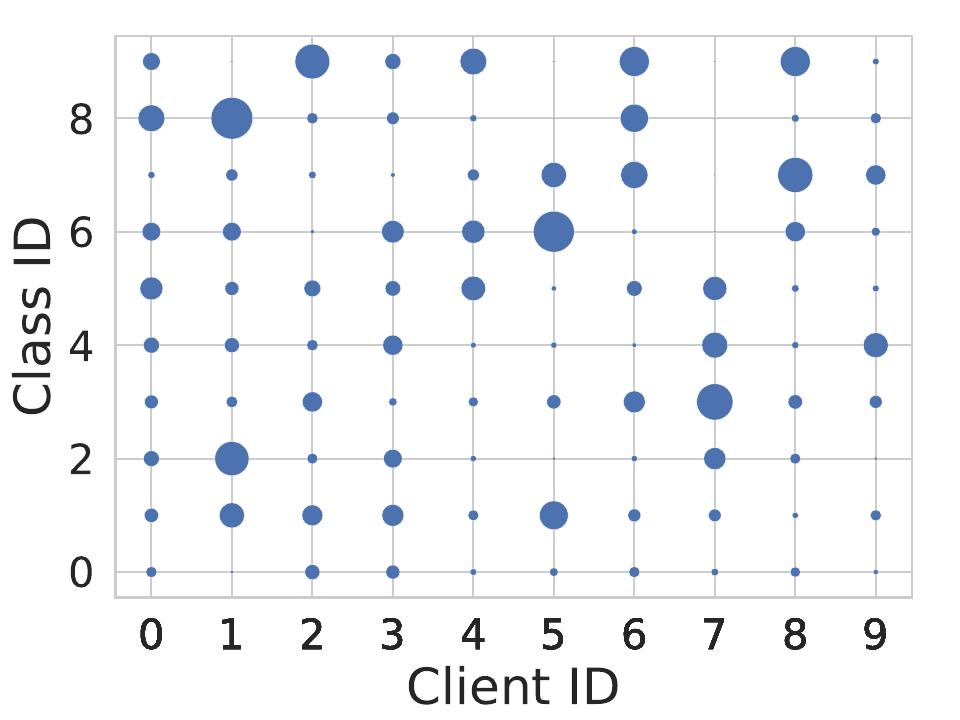}
		\label{fig:dis_d1.0}
	}
 	\vspace{-0.1in}
	\caption{Data distributions of selected clients with different non-IID settings.}
	\label{fig:data_dis}
\end{figure}

\subsubsection{Dataset Settings}\label{sec:exp_setting}
We conducted experiments on five well-known datasets, 
i.e., CIFAR-10, CIFAR-100~\cite{krizhevsky2009learning}, FEMNIST, Shakespeare, and Sent140~\cite{caldas2018leaf}.
To evaluate the performance of FedCross within both
IID and non-IID scenarios, we adopted
the Dirichlet distribution~\cite{hsu2019measuring} denoted by $Dir(\beta)$
to control the heterogeneity settings for datasets 
 CIFAR-10 and CIFAR-100, where 
 a smaller $\beta$ indicates a  higher data heterogeneity
of clients.
For these two datasets, we assumed that there are 100 clients involved in FL. 
To show the quantity 
differences of samples on clients 
within non-IID scenarios for the CIFAR-10 experiment,  
Figure~\ref{fig:data_dis}  shows the data distributions of ten  
clients randomly selected from these 100 clients, where a larger blue dot indicates more samples on the corresponding device. 
%
Unlike CIFAR-10 and CIFAR-100,  the other three datasets (i.e.,  FEMNIST, Shakespeare, and Sent140) are naturally non-IID in terms of  
   data heterogeneity (i.e., number of samples and class imbalance). 
For FEMNIST, Shakespeare, and Sent140, we assumed that there are 180, 128, and 803 clients involved in FL, and each client has more than 100, 5700, and 40 samples, respectively.



\begin{table}[h]
\vspace{-0.05in}
\caption{Comparison between baseline methods and FedCross
}
\vspace{-0.05in}
\label{tab:baseline}
\centering
\scriptsize
\begin{tabular}{|c|c|c|}
\hline
\bf Method & \bf Category & \bf Comm. Overhead\\
\hline\hline
FedAvg & Classic & Low\\
FedProx & Global Control Variable &Low\\
SCAFFOLD & Global Control Variable & High\\
FedGen & Knowledge Distillation & Medium\\
CluSamp & Client Grouping & Low\\
FedCross & Multi-Model Guided & Low \\
\hline
\end{tabular}
\end{table}

\begin{table*}[ht]
\centering
\caption{Test accuracy comparison for both non-IID and  IID scenarios using three DL models}
\label{tab:acc}
\vspace{-0.1 in}
 \scriptsize
\begin{tabular}{|c|c|c|c|c|c|c|c|c|}
\hline
\multirow{2}{*}{\bf Model} & \multirow{2}{*}{\bf Dataset} & \bf Heterogeneity & \multicolumn{6}{c|}{\bf Test Accuracy (\%)} \\
\cline{4-9}
 & & \bf Settings & \bf FedAvg & \bf FedProx &  \bf SCAFFOLD & \bf FedGen & \bf CluSamp & {\bf FedCross} \\
\hline
\hline
\multirow{9}{*}{CNN} & \multirow{4}{*}{CIFAR-10} & $\beta=0.1$& $46.12\pm 2.35$& $47.17\pm 1.65$& $49.12\pm 0.91$ & $ 49.27\pm 0.85 $ & $47.09\pm 0.97$ & ${\bf 55.70\pm 0.74}$\\
                      &  & $\beta=0.5$ & $52.82\pm 0.91$ & $53.59\pm 0.88$ & $54.50\pm 0.44$  & $51.77\pm 0.73$  & $54.00\pm 0.38$ & ${\bf 58.74\pm 0.67}$\\
                      &  & $\beta=1.0$ & $54.78\pm 0.56$ & $54.96\pm 0.60$ & $56.75\pm 0.26$  & $55.38\pm 0.66$  & $55.82\pm 0.73$ & ${\bf 62.16\pm 0.42}$\\
                      &  & $IID$ & $57.64\pm 0.22$ & $58.34\pm 0.15$ & $59.98\pm 0.22$  & $58.71\pm 0.19$ & $57.32\pm 0.21$ & ${\bf 62.97\pm 0.22}$\\
\cline{2-9}
& \multirow{4}{*}{CIFAR-100} & $\beta=0.1$ & $28.37\pm 1.10$ & $28.11\pm 1.03$ & $30.32\pm 1.05$ & $28.18\pm 0.58$  & $28.63\pm 0.63$   & ${\bf 32.53\pm 0.45}$\\
                     &  & $\beta=0.5$     & $30.01\pm 0.56$ & $32.16\pm 0.50$ & $33.49\pm 0.73$  & $29.55\pm 0.41$  & $33.04\pm 0.41$ & ${\bf 36.87\pm 0.24}$\\
                     &  & $\beta=1.0$      & $32.34\pm 0.65$ & $32.78\pm 0.13$ & $34.95 \pm 0.58$  & $31.88\pm 0.65$  & $32.92\pm 0.31$ & ${\bf 37.65\pm 0.36}$\\
                     &  & $IID$ & $32.98\pm 0.20$ & $33.39\pm 0.25$ & $35.11\pm 0.23$   & $32.43\pm 0.20$ & $34.97\pm 0.24$ & ${\bf 38.42\pm 0.18}$\\
\cline{2-9}
& \multirow{1}{*}{FEMNIST} & $-$ & $81.67\pm 0.36$ & $82.10\pm 0.61$ & $81.65\pm 0.21$  & $81.95 \pm 0.36$  & $80.80\pm 0.40$  &  ${\bf 83.49\pm 0.18}$ \\
\hline
\hline
\multirow{10}{*}{ResNet-20} & \multirow{4}{*}{CIFAR-10} & $\beta=0.1$& $45.11\pm 2.13$& $45.45\pm 3.42$& $50.46\pm 1.76$  & $42.71\pm 3.48$  & $44.87\pm 1.65$ & ${\bf 53.79\pm 2.91}$\\
                      &  & $\beta=0.5$ & $60.56\pm 0.95$ & $59.52\pm 0.74$ & $58.85\pm 0.85$ & $60.29\pm 0.68$  & $59.55\pm 1.00$ & ${\bf 69.38\pm 0.30}$\\
                      &  & $\beta=1.0$ & $62.99\pm 0.62$ & $61.47\pm 0.66$ & $61.63\pm 0.78$  & $63.81\pm 0.33$  & $63.32\pm 0.71$ & ${\bf 71.59\pm 0.31}$\\
                      &  & $IID$ & $67.12\pm 0.27$ & $66.06\pm 0.22$ & $65.20\pm 0.27$  & $65.89\pm 0.17$ & $65.62\pm 0.23$ & ${\bf 75.01\pm 0.09}$\\
\cline{2-9}
& \multirow{4}{*}{CIFAR-100} & $\beta=0.1$ & $31.90\pm 1.16$ & $33.00\pm 1.21$ & $35.71\pm 0.62$ &  $32.40\pm 1.45$  & $34.34\pm 0.52$   & ${\bf 39.40\pm 1.43}$\\
                     &  & $\beta=0.5$     & $42.45\pm 0.53$ & $42.83\pm 0.54$ & $42.33\pm 1.23$ & $42.72\pm 0.32$ & $42.07\pm 0.39$ & ${\bf 50.39\pm 0.24}$\\
                     &  & $\beta=1.0$      & $44.22\pm 0.36$ & $44.35\pm 0.36$ & $43.28\pm 0.61$  & $44.75\pm 0.57$  & $43.29\pm 0.41$ & ${\bf 53.09\pm 0.29}$\\
                     &  & $IID$ & $44.42\pm 0.18$ & $45.16\pm 0.24$ & $44.37\pm 0.19$   & $45.21\pm 0.19$ & $43.59\pm 0.24$ & ${\bf 54.07\pm 0.19}$\\
\cline{2-9}
& \multirow{1}{*}{FEMNIST} & $-$ & $78.47\pm 0.40$ & $79.74\pm 0.54$ & $76.14\pm 0.90$ & $79.56\pm 0.34$  & $79.28\pm 0.42$  &  ${\bf 80.93\pm 0.52}$ \\
\hline
\hline
\multirow{9}{*}{VGG-16} & \multirow{4}{*}{CIFAR-10} & $\beta=0.1$& $63.79\pm 3.90$& $63.35\pm 4.31$& $64.18\pm 3.86$ &  $66.52\pm 1.46$  & $66.91\pm 1.83$ & ${\bf 76.07\pm 1.09}$\\
                      &  & $\beta=0.5$ & $78.14\pm 0.67$ & $77.70\pm 0.45$ & $76.22\pm 1.37$ & $78.9\pm 0.39$  & $78.82\pm 0.40$ & ${\bf 84.39\pm 0.48}$\\
                      &  & $\beta=1.0$ & $78.55\pm 0.21$ & $79.10\pm 0.28$ & $76.99\pm 1.01$ & $79.75\pm 0.26$ & $80.00\pm 0.37$ & ${\bf 85.74\pm 0.21}$\\
                      &  & $IID$ & $80.02\pm 0.05$ & $80.77\pm 0.22$ & $78.80\pm 0.07$ & $80.00\pm 0.27$  & $80.96\pm 0.12$ & ${\bf 87.33\pm 0.11}$\\
\cline{2-9}
& \multirow{4}{*}{CIFAR-100} & $\beta=0.1$ & $46.60\pm 1.45$ & $45.88\pm 3.35$ & $45.79\pm 1.77$ & $49.04\pm 0.63$  & $48.04\pm 1.76$   & ${\bf 54.46\pm 0.70}$\\
                     &  & $\beta=0.5$     & $55.86\pm 0.64$ & $55.79\pm 0.56$ & $55.30\pm 0.61$ & $56.40\pm 0.37$  & $56.23\pm 0.34$ & ${\bf 64.01\pm 0.24}$\\
                     &  & $\beta=1.0$      & $57.55\pm 0.51$ & $57.40\pm 0.32$ & $55.43\pm 0.45$ & $57.15\pm 0.27$  & $57.95\pm 0.35$ & ${\bf 67.09\pm 0.31}$\\
                     &  & $IID$ & $58.30\pm 0.23$ & $58.49\pm 0.11$ & $56.51\pm 0.08$ & $57.62\pm 0.18$ & $58.14\pm 0.20$ & ${\bf 70.81\pm 0.07}$\\
\cline{2-9}
& \multirow{1}{*}{FEMNIST} & $-$ & $84.22\pm 0.46$ & $83.98\pm 0.48$ & $82.65\pm 0.74$ & $84.69\pm0.28$ & $84.32\pm 0.36$  &  ${\bf 85.75\pm 0.45}$\\
\hline
\hline

\multirow{2}{*}{LSTM}& Shakespeare & $-$ & $52.08\pm 0.29$ & $52.53\pm 0.23$& $48.94\pm 0.18$ & $ 53.87\pm 0.13 $ & $49.74\pm 0.74$ & ${\bf 54.81\pm 0.07}$\\
\cline{2-9}
& Sent140 & $-$ & $69.36\pm 0.20$& $68.63\pm 0.20$& $59.61\pm 0.06$ & $ 69.32\pm 0.13$ & $69.19\pm 0.14$ & ${\bf 71.33\pm 0.12}$\\
\hline

\end{tabular}
\vspace{-0.2in}
\end{table*}

\subsubsection{Baseline Methods and Their  Settings}
We compared FedCross with five baseline methods, including 
 the classic FedAvg and four 
 state-of-the-art FL optimization methods (i.e., FedProx, SCAFFOLD, FedGen, and CluSamp).
 Table~\ref{tab:baseline} compares FedCross with all the baseline methods from the perspectives of categories and communication overheads, where the baselines cover all the three FL optimization categories introduced in Section~\ref{sec:relwork}. 
 Note that, as a novel multi-model guided FL method, FedCross
 does not belong to any of the three existing categories.
 The following presents their settings.
\begin{itemize}[leftmargin=*]
    \item \textbf{FedAvg}~\cite{mcmahan2017communication} is the most classic
    one-to-multi FL framework, wherein each FL training round the cloud server dispatches a global model to selected clients for FL training and aggregates their trained local models averagely to update the global model.
    
    \item \textbf{FedProx}~\cite{li2020federated} is a global control variable-based 
    FL framework
    influenced by the hyper-parameter $\mu$, where
    $\mu$ controls the weight of its proximal term.
    We set the best  $\mu$ values for CIFAR-10, CIFAR-100, and FEMNIST to 0.01, 0.001, and 0.1, respectively. All these values are explored from the set \{0.001, 0.01, 0.1, 1.0\}.
    
    \item \textbf{SCAFFOLD}~\cite{karimireddy2020scaffold} is  a global control variable-based  FL framework, where the cloud server 
      dispatches the variable with
     the same size as the   model to guide local training in each  training
     round.
    
    \item \textbf{FedGen}~\cite{zhu2021data} is a KD-based method, which includes a built-in generator for proxy dataset generation. 
    The subsequent experiments used the same settings as in \cite{zhu2021data}.
    
    \item \textbf{CluSamp}~\cite{fraboni2021clustered} is a client grouping-based method. We select the model gradient similarity as the criteria for client grouping rather than the sample size. This is because directly exposing the distribution of data may increase the risk of privacy exposure. Furthermore, it may not be possible to directly obtain data distribution in real scenarios.
\end{itemize}

We implemented all  FL methods on top of our own unified FL framework. For the baselines  FedGen and CluSamp, we re-used the open source code from \cite{fedgencode} and \cite{clusampcode}, respectively. For the baselines  FedProx and SCAFFOLD, we re-implemented them according to their original papers \cite{li2020federated,karimireddy2020scaffold}.

\subsubsection{Model Settings}
We investigated
three well-known models, i.e., CNN, ResNet-20~\cite{he2016deep}, VGG-16~\cite{simonyan2014very}.
The CNN model was obtained from FedAvg~\cite{mcmahan2017communication}, consisting of two convolutional and fully-connected layers. ResNet-20 and VGG-16 models were obtained from the official library \cite{models}. 

\begin{figure}[h]
	\vspace{-0.2in}
	\centering
	\subfigure[FedAvg with $\beta=0.1$]{
		\centering
		\includegraphics[width=0.16\textwidth]{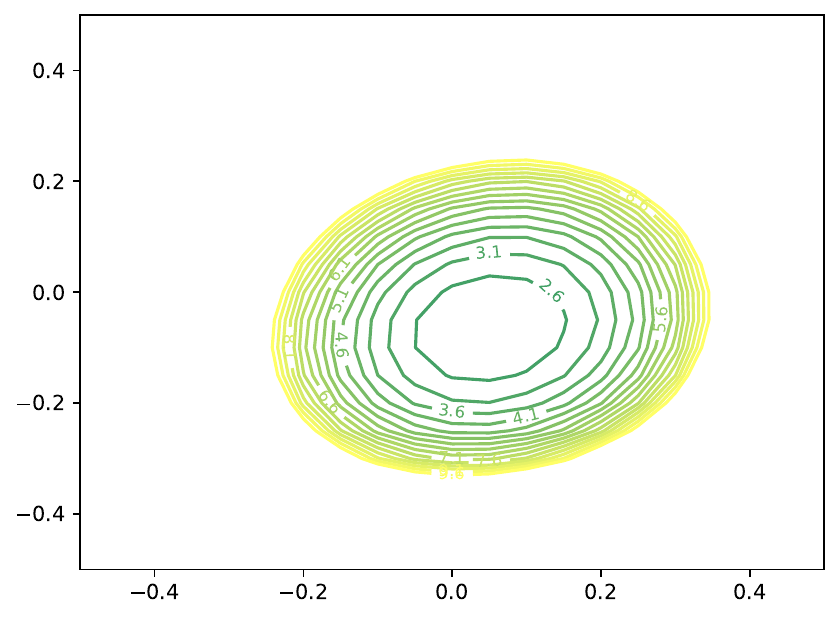}
		\label{fig:fedavg_landscape_d0.1}
	}\hspace{0.1in} \vspace{-0.05in}
	\subfigure[FedCross with $\beta=0.1$]{
		\centering
		\includegraphics[width=0.16\textwidth]{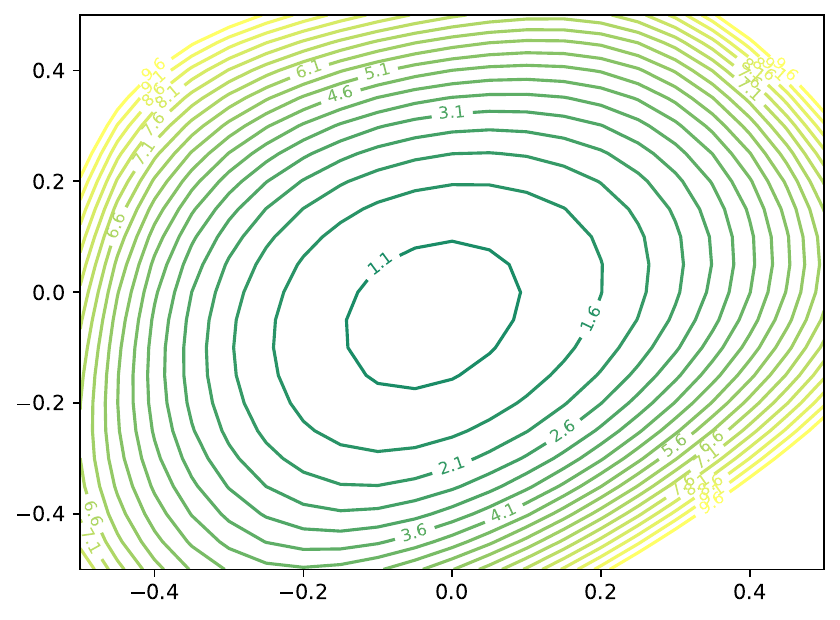}
  \label{fig:fedcross_landscape_d0.1}
        }
        \subfigure[FedAvg with $IID$]{
		\centering
		\includegraphics[width=0.16\textwidth]{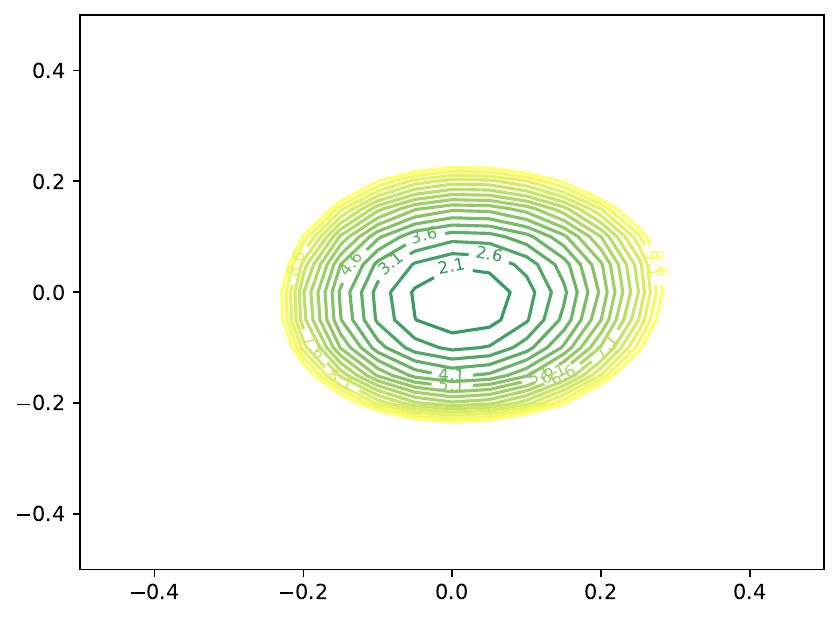}
		\label{fig:fedavg_landscape_iid}
	}\hspace{0.1in}
	\subfigure[FedCross with $IID$]{
		\centering
		\includegraphics[width=0.16\textwidth]{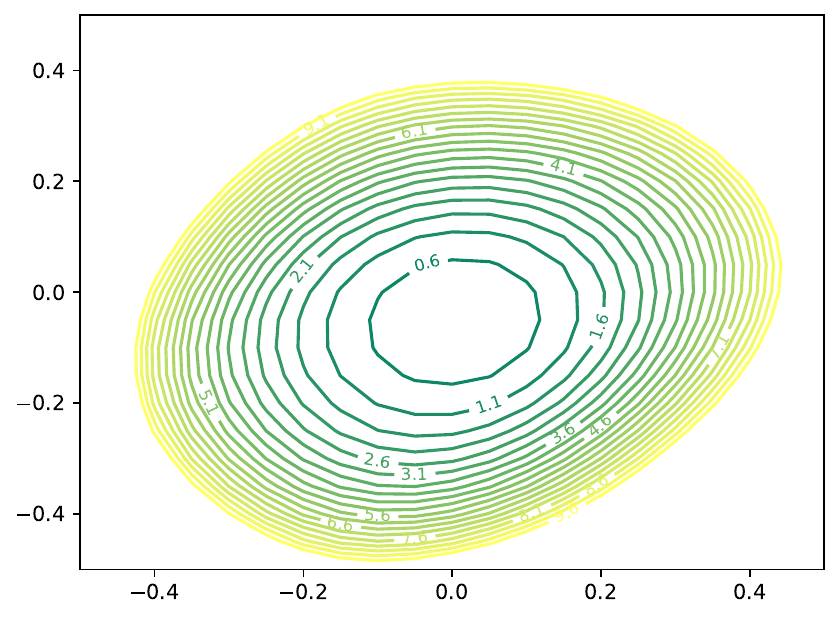}
  \label{fig:fedcross_landscape_iid}
        }
 	\vspace{-0.05in}
	\caption{Comparison between loss landscapes of FedAvg and FedCross.}
	\label{fig:val_landscape}
 	\vspace{-0.1in}
\end{figure}

\subsection{Motivation Validation (RQ1)}\label{sec:exp_val}
To validate whether a global model trained by FedCross can converge into a flatter valley than FedAvg, we checked four models for ResNet-20 that are trained using both  FedAvg and FedCross on the CIFAR-10 dataset with $\beta=0.1$ and IID scenarios, respectively.
Since it is hard to 
draw the landscapes of all the
involved clients together,
Figure~\ref{fig:val_landscape}
only shows the loss landscapes of the obtained global models on top of their corresponding whole datasets.  From this figure, we can observe that the global models trained by FedAvg are located in sharper areas than those obtained by FedCross.
This implicitly
reflects the fact that all the clients converge into nearby flat optimal solution areas, which is consistent with our observation in Figure~\ref{fig:motivation}.
In other words, from the perspective of loss landscapes, FedCross can train a more generalized global model than that trained by FedAvg.

\begin{figure*}[ht]
	\centering
	\subfigure[CNN with $\beta=0.1$]{
		\centering
		\includegraphics[width=0.18\textwidth]{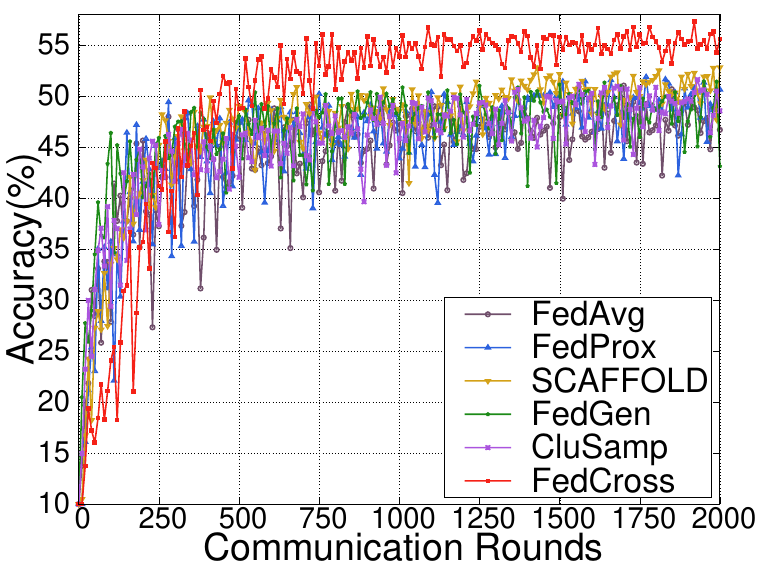}	
  \label{fig:cnn-0.1}
	}\vspace{-0.02 in}
	\subfigure[CNN  with $\beta=0.5$]{
		\centering
		\includegraphics[width=0.18\textwidth]{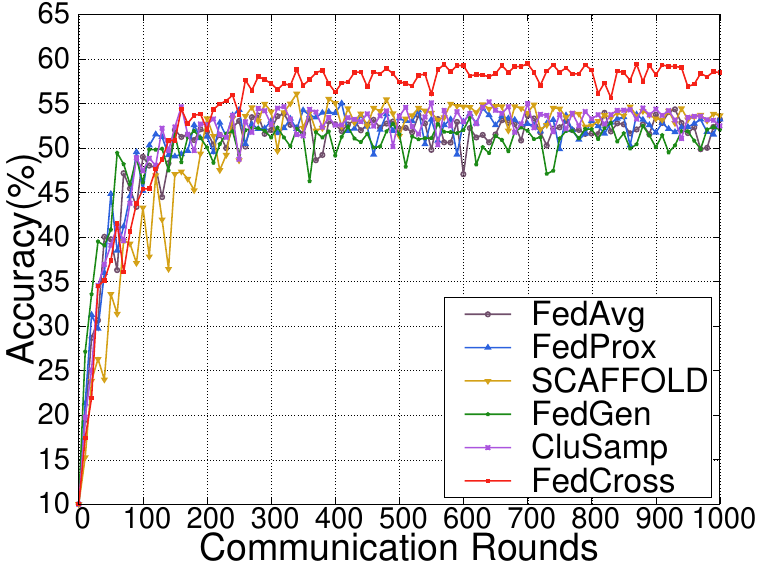}
		\label{fig:cnn-0.5}
	}
	\subfigure[CNN with $\beta=1.0$]{
		\centering
		\includegraphics[width=0.18\textwidth]{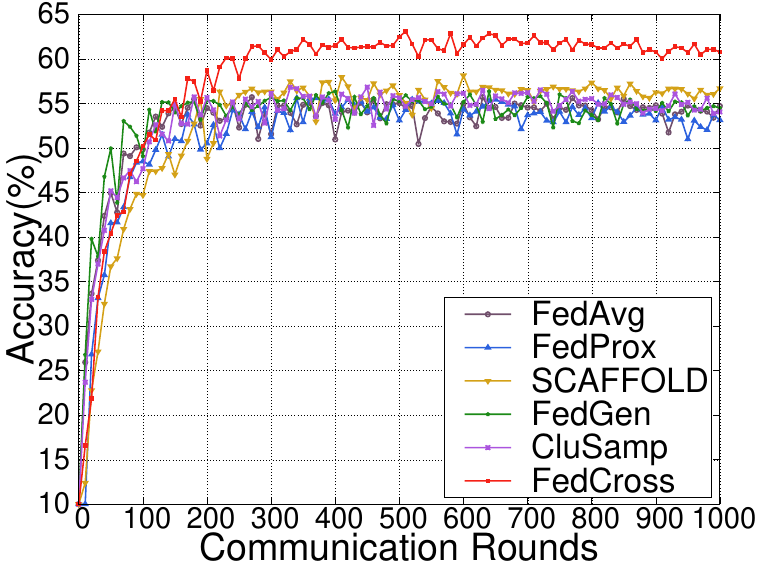}
		\label{fig:cnn-1.0}
	}\vspace{-0.02 in}
	\subfigure[CNN with IID]{
		\centering
		\includegraphics[width=0.18\textwidth]{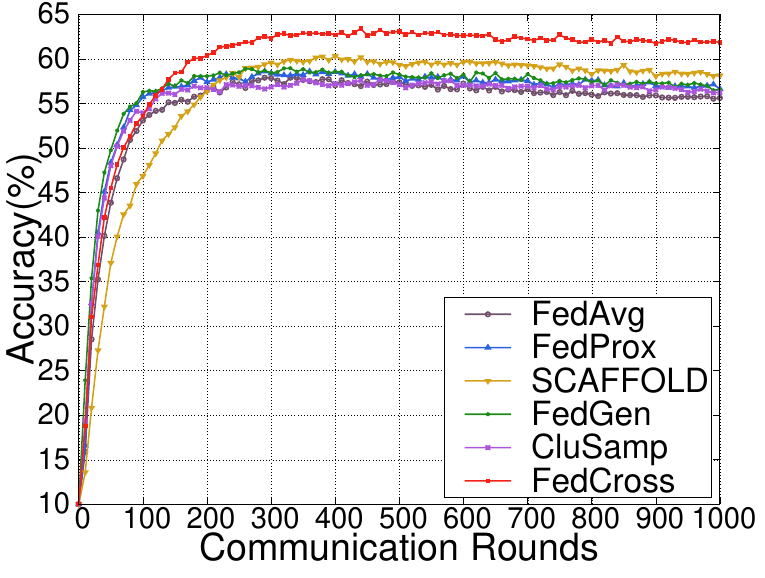}
		\label{fig:cnn-iid}
	}\vspace{-0.02 in}
	
	\subfigure[ResNet-20 with $\beta=0.1$]{
		\centering
		\includegraphics[width=0.18\textwidth]{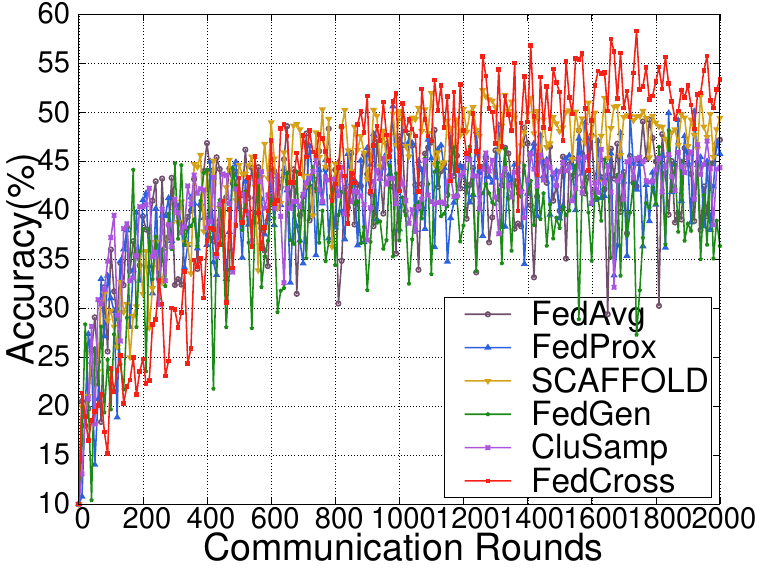}
		\label{fig:resnet-0.1}
	}
	\subfigure[ResNet-20  with $\beta=0.5$]{
		\centering
		\includegraphics[width=0.18\textwidth]{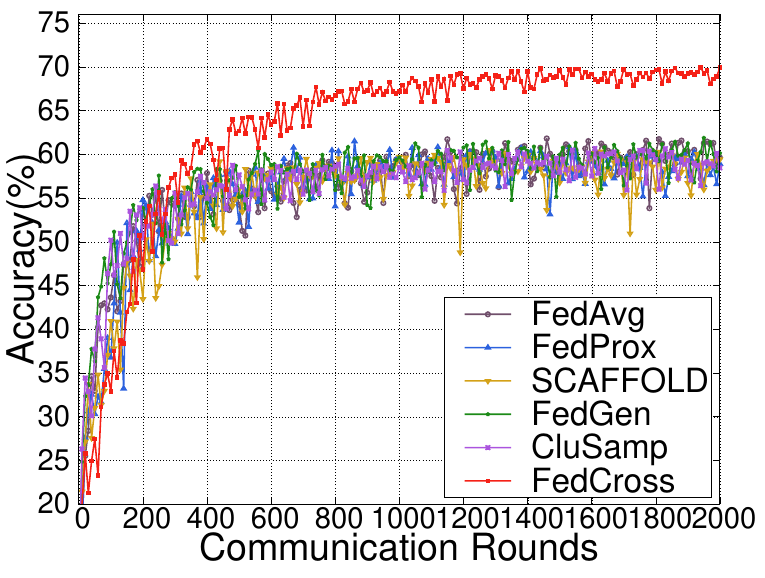}
		\label{fig:resnet-0.5}
	}\vspace{-0.02 in}
	\subfigure[ResNet-20 with $\beta=1.0$]{
		\centering
		\includegraphics[width=0.18\textwidth]{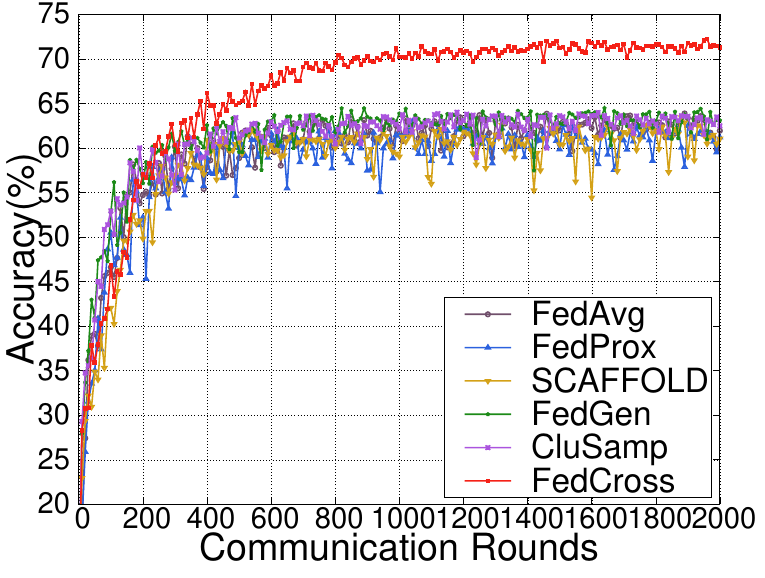}
		\label{fig:resnet-1.0}
	}\vspace{-0.02 in}
	\subfigure[ResNet-20 with IID]{
		\centering
		\includegraphics[width=0.18\textwidth]{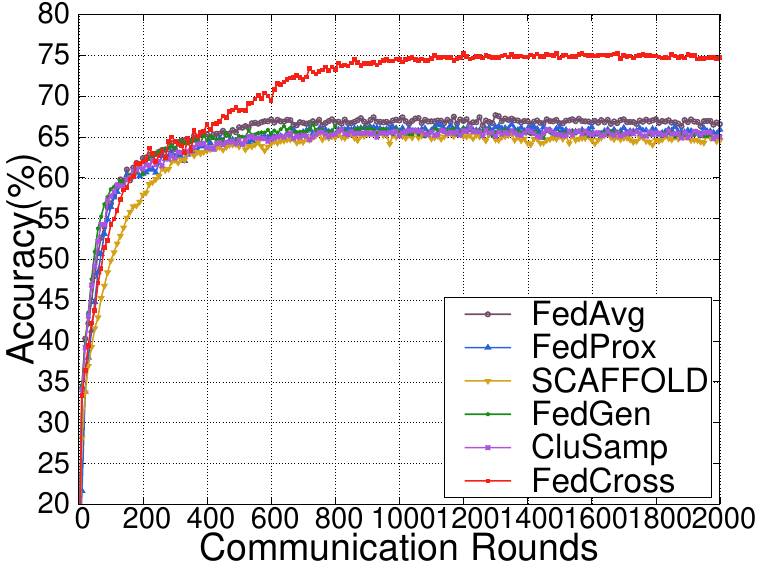}
		\label{fig:resnet-iid}
	}\vspace{-0.02 in}
	
	\subfigure[VGG-16 with $\beta=0.1$]{
		\centering
		\includegraphics[width=0.18\textwidth]{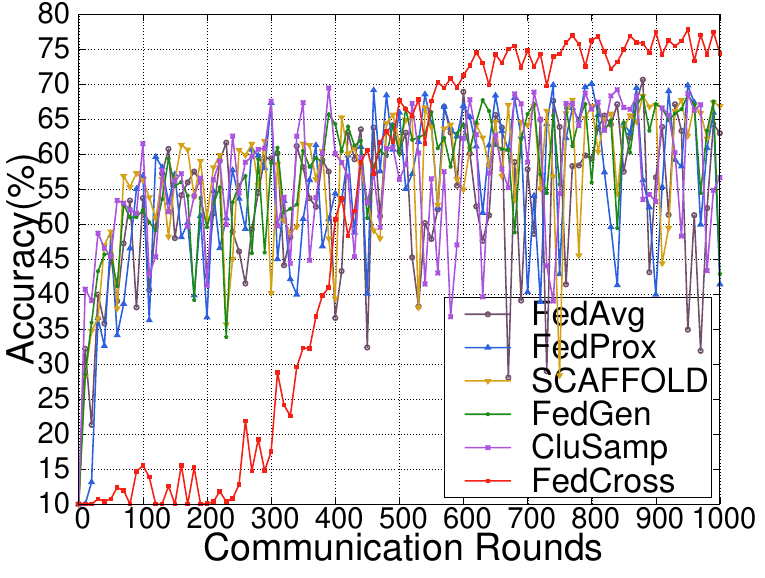}
		\label{fig:vgg-0.1}
	}
	\subfigure[VGG-16  with $\beta=0.5$]{
		\centering
		\includegraphics[width=0.18\textwidth]{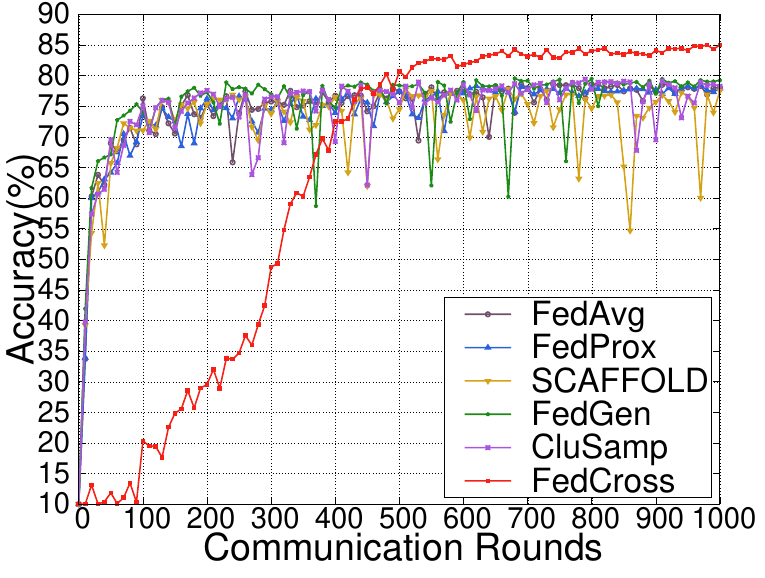}
		\label{fig:vgg-0.5}
	}
	\subfigure[VGG-16 with $\beta=1.0$]{
		\centering
		\includegraphics[width=0.18\textwidth]{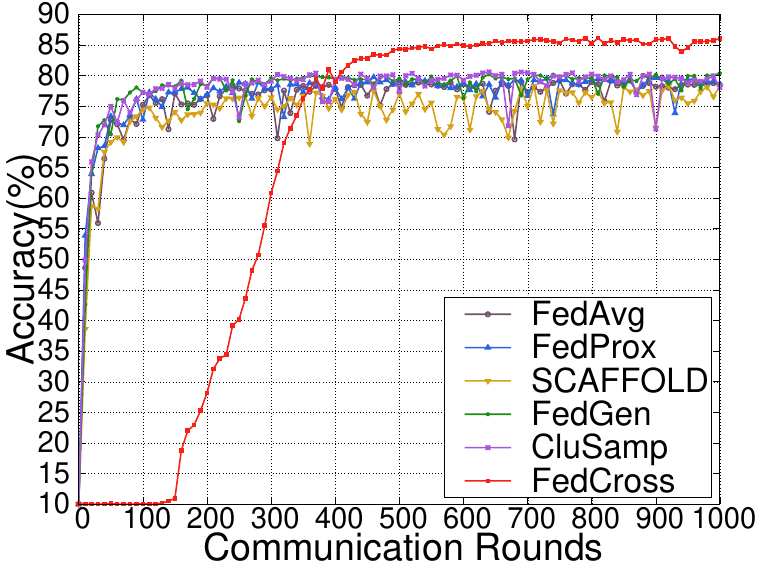}
		\label{fig:vgg-1.0}
	}
	\subfigure[VGG-16 with IID]{
		\centering
		\includegraphics[width=0.18\textwidth]{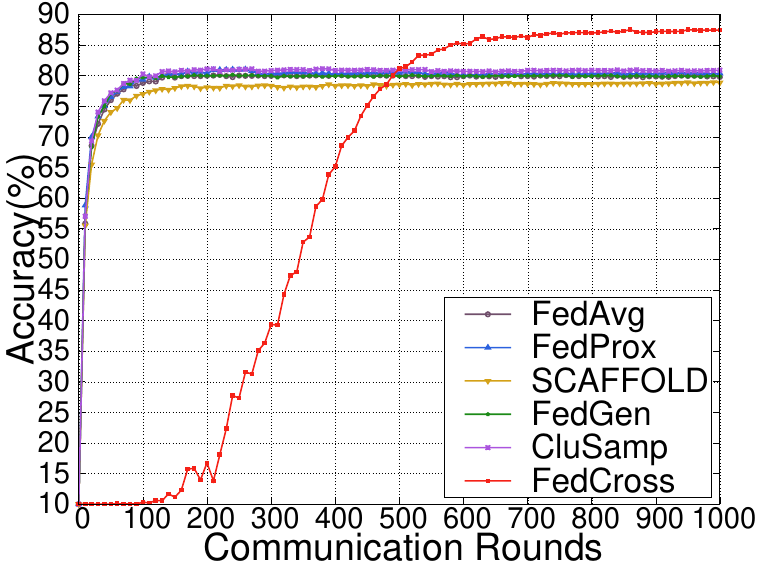}
		\label{fig:vgg-iid}
	}
\vspace{-0.1in}
	\caption{Learning curves of different FL methods on CIFAR-10 dataset.}
  \vspace{-0.2in}
	\label{fig:accuacy}
\end{figure*}

\subsection{Performance Comparison (RQ2)}\label{sec:exp_acc}

To show the superiority of FedCross, we compared it with the five baselines. 
For datasets CIFAR-10 and CIFAR-100, we considered one IID and three non-IID scenarios (with $\beta=0.1,0.5,1.0$, respectively).

\subsubsection{Comparison of Inference Accuracy}\label{sec:exp_acc}
Table~\ref{tab:acc} presents the classification 
accuracy results for FedCross and all the five baselines on three datasets, 
where both 
IID and non-IID scenarios are all investigated.
Note that, in the third column, we use $\beta$ to control 
the heterogeneity settings for datasets CIFAR-10 and CIFAR-100 
based on Diricht distribution $Dir$. 
Note that for all the baselines, we set the numbers of FL training rounds to 2000, 2000, and 1000 when using the CNN, ResNet-20, and VGG-16 models, respectively. We set the number of FL training rounds to 1000 for the ShakeSpeare dataset and 3000 for the Sent140 dataset.
From this table, 
we can observe that FedCross achieves the highest accuracy for all different settings. 
%
For example, when using the VGG-16 model on CIFAR-10,
FedCross outperforms the best baseline counterparts by 9.16\% and 6.37\% within 
IID  and non-IID ($\beta=0.1$) scenarios, respectively.
Note that, by merely replacing
the one-to-multi training scheme in the FedAvg framework with  our 
proposed multi-to-multi training scheme, 
the classification performance of FedCross can be improved 
dramatically.
One may argue that the classification performance improvements 
made by FedCross for FEMNIST are not as significant as the ones obtained for  
datasets CIFAR-10 and CIFAR-100. This is mainly because 
the data samples are simpler than the ones in datasets CIFAR-10 and CIFAR-100, 
where even FedAvg can achieve near-optimal classification performance. 
Moreover, 
we can observe that FedCross achieves the best performance on the two text datasets, i.e., ShakeSpeare and Sent140.



\begin{figure*}[ht]
	\centering
	\subfigure[$K=5$]{
		\centering
		\hspace{-0.1in}
  \includegraphics[width=0.18\textwidth]{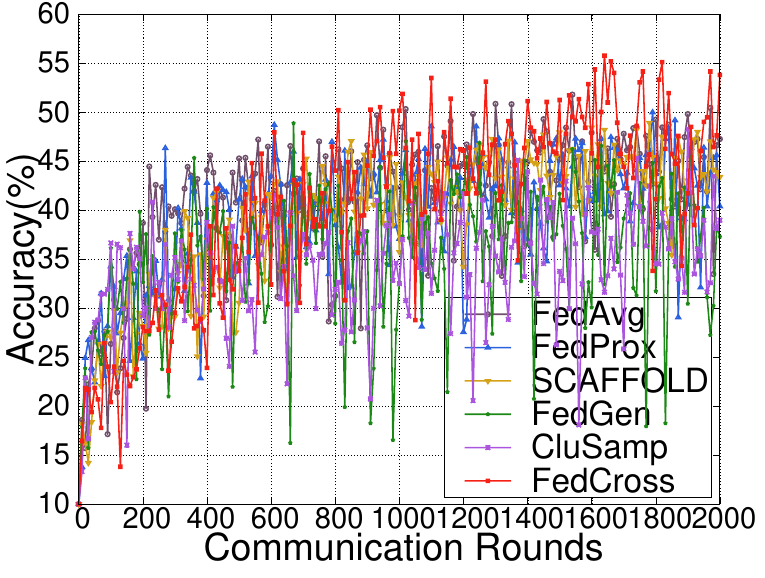}
		\label{fig:frac_0.05}
	} \hspace{-0.15in}
        \subfigure[$K=10$]{
		\centering
		\includegraphics[width=0.18\textwidth]{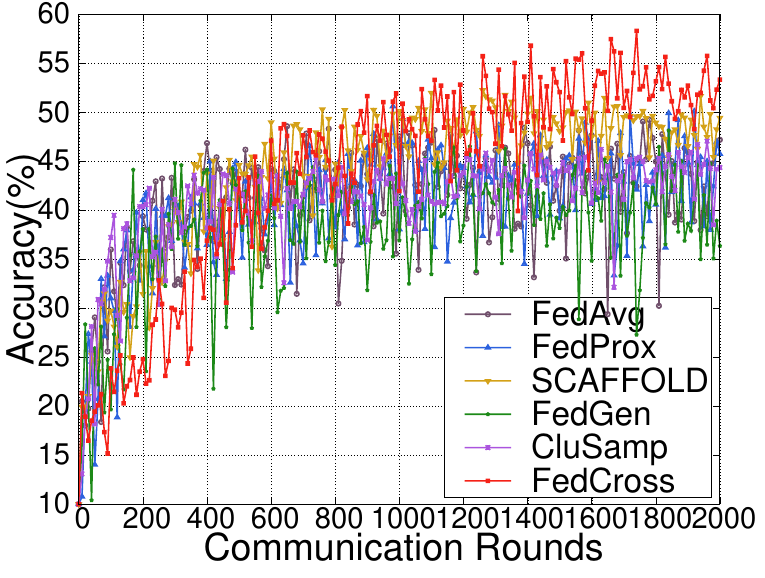}
		\label{fig:frac_0.1}
	} \hspace{-0.15in}
	\subfigure[$K=20$]{
		\centering
		\includegraphics[width=0.18\textwidth]{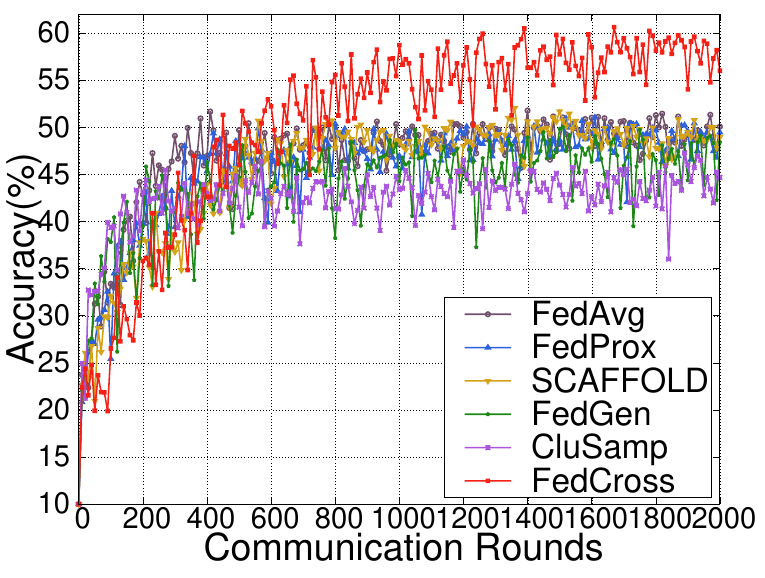}
		\label{fig:frac_0.2}
	} \hspace{-0.15in}
	\subfigure[$K=50$]{
		\centering
		\includegraphics[width=0.18\textwidth]{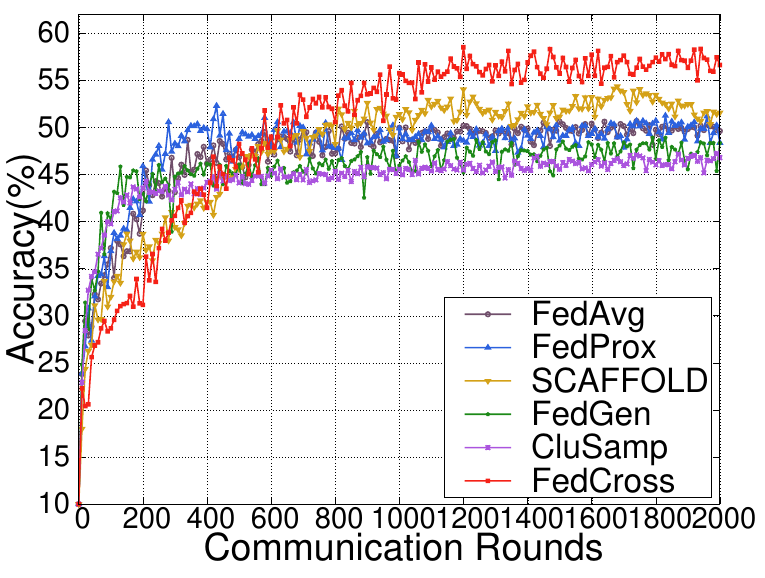}
		\label{fig:frac_0.5}
	} \hspace{-0.15in}
	\subfigure[$K=100$]{
		\centering
		\includegraphics[width=0.18\textwidth]{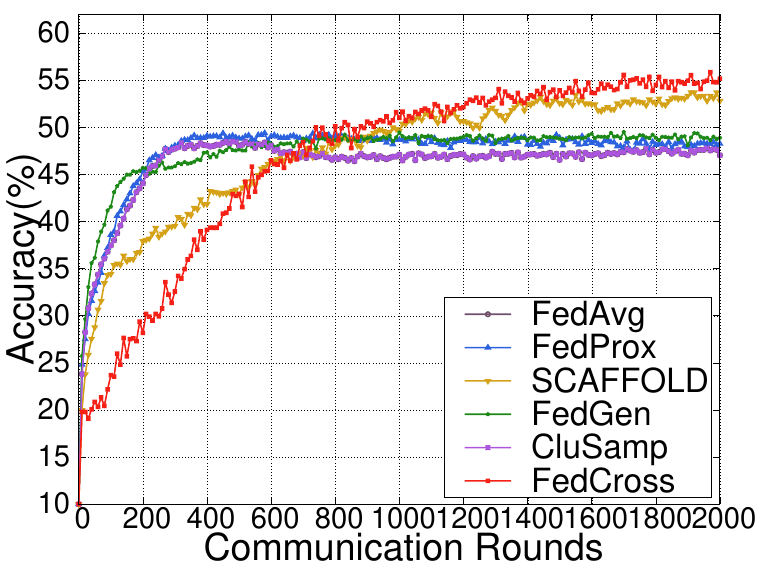}
		\label{fig:frac_1.0}
	}
 \vspace{-0.05in}
	\caption{Learning curves of different ResNet-20-based FL methods for different number of activated clients on CIFAR-10 dataset with $\alpha=0.1$.}
   \vspace{-0.1in}
	\label{fig:frac}
\end{figure*}

\begin{figure*}[ht]
	\centering
	\subfigure[$|C|=50$]{
		\centering
		\hspace{-0.1in}
  \includegraphics[width=0.18\textwidth]{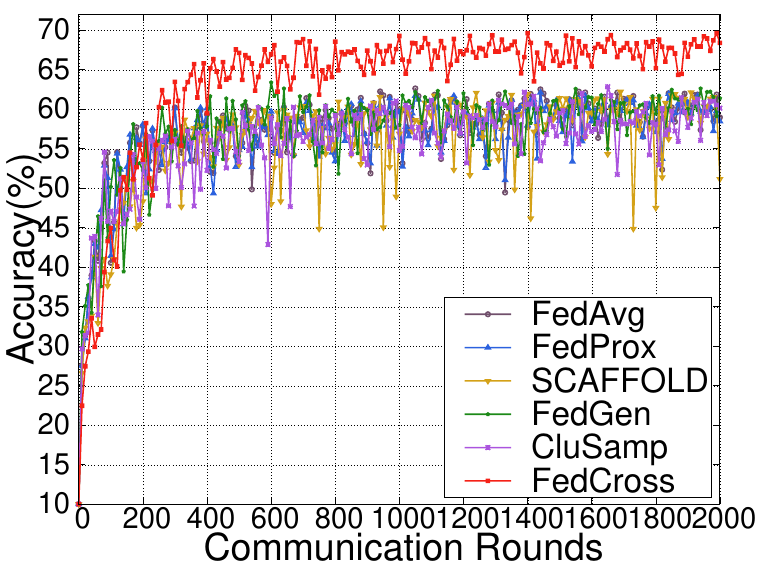}
		\label{fig:frac_0.05}
	} \hspace{-0.15in}
        \subfigure[$|C|=100$]{
		\centering
		\includegraphics[width=0.18\textwidth]{fig/accuracy_resnet_cifar_d0.5.pdf}
		\label{fig:frac_0.1}
	} \hspace{-0.15in}
	\subfigure[$|C|=200$]{
		\centering
		\includegraphics[width=0.18\textwidth]{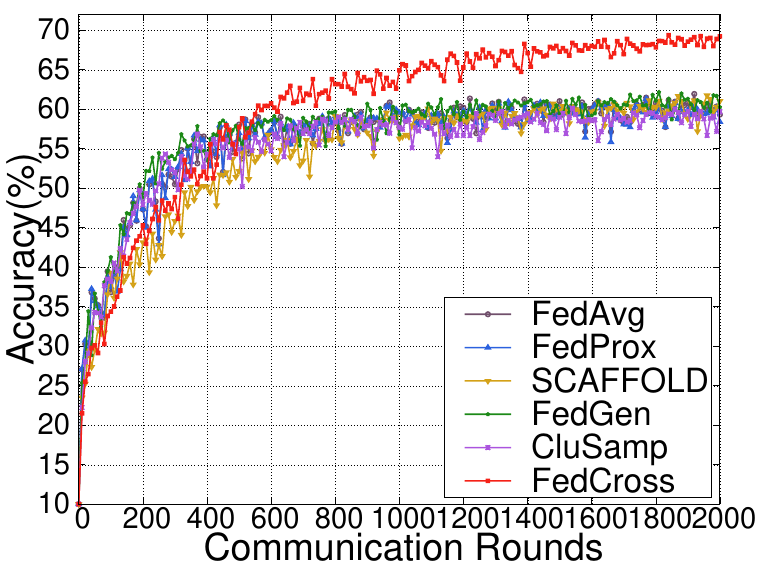}
		\label{fig:frac_0.2}
	} \hspace{-0.15in}
	\subfigure[$|C|=500$]{
		\centering
		\includegraphics[width=0.18\textwidth]{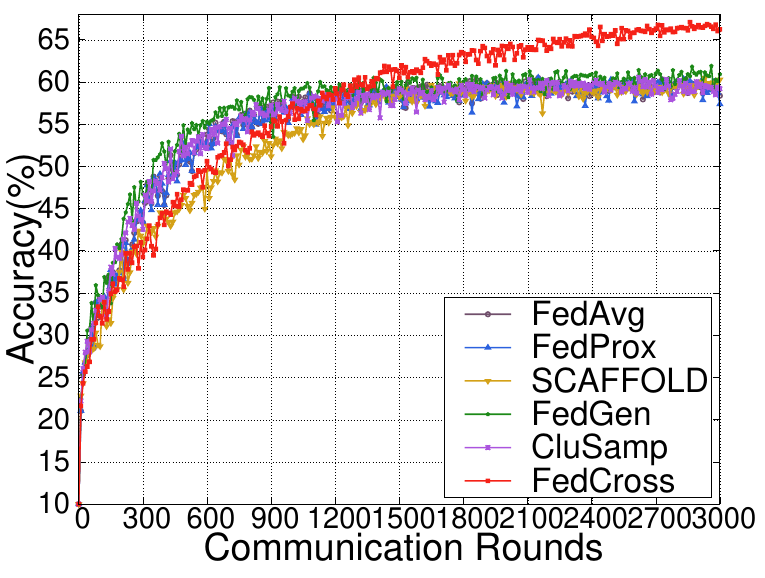}
		\label{fig:frac_0.5}
	} \hspace{-0.15in}
	\subfigure[$|C|=1000$]{
		\centering
		\includegraphics[width=0.18\textwidth]{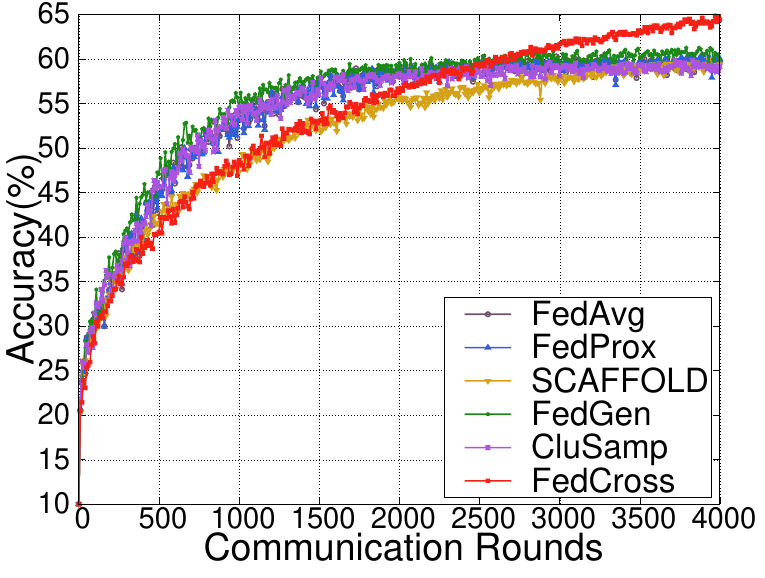}
		\label{fig:frac_1.0}
	}
 \vspace{-0.05in}
	\caption{Learning curves of different ResNet-20-based FL methods for different number of clients on CIFAR-10 dataset with $\alpha=0.5$.}
   \vspace{-0.15in}
	\label{fig:client_num}
\end{figure*}

\subsubsection{Comparison of  Convergence Rate}
Figure~\ref{fig:accuacy} shows the convergence trends 
of all the FL methods (including five baselines and FedCross) on the CIFAR-10 dataset, where Figures~\ref{fig:cnn-0.1}-\ref{fig:cnn-iid} use CNN model, Figures~\ref{fig:resnet-0.1}-\ref{fig:resnet-iid} use ResNet-20 model, and Figures~\ref{fig:vgg-0.1}-\ref{fig:vgg-iid} use VGG-16 model.
FedCross does not generate global models along with the FL training process. 
To enable the classification accuracy comparison between FedCross and the baselines,
we additionally generated one pseudo-global model based on the middleware models 
in each round of FL training, and adopted this global model to derive the test accuracy
information.

From Figure~\ref{fig:accuacy}, we can find that FedCross consistently achieves the highest accuracy performance of  the six 
FL methods in both non-IID and IID scenarios. 
Furthermore, we can observe that FedCross converges with  much smaller fluctuations
for all the investigated models and data settings. 
This is mainly because FedCross uses a multi-to-multi training scheme
based on our proposed 
 multi-model cross-aggregation, leading to
 the fine-grained training of the global model. 
 Due to mitigated gradient divergence during local training  and the available access of data across 
 clients, FedCross can achieve the highest test accuracy results while lowering the 
 risk of stuck-at-local-training. 
As shown in Figures~\ref{fig:vgg-0.1}-\ref{fig:vgg-iid}, at the beginning of 
 FL training, FedCross lags behind the five baselines.
This phenomenon is mainly because VGG-16 is a connection-intensive model with more than 130 million parameters, while  ResNet-20 only has about 30 million parameters. 
Since  VGG-16 is much larger than ResNet-20, it has a smaller performance 
acceleration than 
ResNet-20 at the early phase of FL training.



\subsubsection{Comparison of Communication Overhead}

For FedAvg,  each training round involves 
the dispatching of $K$ models  and the upload of $K$ models in total, 
where $K$ is the number of selected clients. 
Although FedCross uses multiple models for FL training, it does not increase communication overhead than FedAvg.
For FedCross, each participant 
client  in local 
training receives only one model and uploads its
trained version. 
Therefore,  each training round  of FedCross   
needs a communication of 
$2K$ models, which is the same as FedAvg.
For FedProx and CluSamp, since their communication does not involve parameters other than models, their communication overhead is the same as FedAvg.
For SCAFFOLD, it needs $2K$ models plus $2K$ global control variables in each FL training round,
since
the cloud server dispatches a global control variable to $K$ clients and each client uploads global control variables to the cloud server in each round of FL training.
For FedGen, since the cloud server dispatches an additional built-in generator to $K$ clients in each FL training round, the communication overhead of FedGen is $2K$ models plus $K$ generators.
Based on the above analysis, we can find that FedCross 
requires the least communication overhead in each FL training round.
Note that, as shown in Figure~\ref{fig:accuacy}, although FedCross needs more rounds to achieve its best accuracy, for the highest accuracy that can be achieved by some FL methods, FedCross uses much fewer training rounds than the counterpart. This again shows the communication savings obtained by  FedCross.


\subsection{Compatibility Analysis (RQ3)}
\subsubsection{Impacts of Client Data Distributions}


For the same dataset, although FedCross 
can alleviate the performance degradation caused by various 
data heterogeneity factors, 
compared with their IID counterpart,
the non-IID scenarios still lead to worse
classification performance, especially when $\beta$ is small. 
Furthermore, we find that in non-IID scenarios, FedCross requires more FL training rounds to converge. 
Note that all the above phenomena are also applicable to all the baselines.
In other words, the training in non-IID scenarios is more difficult 
than the training in IID scenarios.
From Table~\ref{tab:acc} and Figure~\ref{fig:accuacy}, we can find that 
FedCross achieves the best performance for both IID and non-IID scenarios. 



\subsubsection{Impacts of Datasets}

From Table~\ref{tab:acc}, we can observe that for all
the three datasets, FedCross can significantly improve
the classification performance compared with baselines,
especially for complex datasets. As an example shown
in Figure~\ref{fig:accuacy}, we can observe  that 
FedCross benefits CIFAR-10 and CIFAR-100 more than FEMNIST.

\subsubsection{Impacts of Models}
From Figure~\ref{fig:accuacy}, we can observe that for the same dataset but different underlying DNN models, FedCross can still achieve the best classification performance. 
When adopting a model with a larger volume of parameters, although
 the convergence of FedCross may 
be slower than the baselines at the beginning of training, we can observe that 
FedCross can achieve much better classification accuracy at the end of training. 
Meanwhile, to achieve the best possible classification accuracy, FedCross uses much
fewer training rounds. To accelerate the convergence of FedCross at the beginning of training, we 
proposed two training acceleration methods. Please refer to Section 
\ref{sec:acceleration} for more details.

\subsubsection{Impacts of Activated Clients}

Figure~\ref{fig:frac} compares FedCross with five baselines on the CIFAR-10 dataset using the ResNet-20 model within a non-IID scenario ($\beta=0.1$), where the number of activated clients investigated in subfigures are 5, 10, 20, 50, and 100, respectively.
From Figure~\ref{fig:frac}, we can observe that FedCross can achieve the best results for all the cases. 
When $K<20$, the maximal classification
accuracy increases along with  the increasing 
number of activated clients. 
However, when $K\ge 20$, the impact of the increasing number of activated clients is negligible. 
Moreover, we can  find that 
the convergence becomes smoother when more activated clients are involved in the FL training. 

\subsubsection{Impacts of the Total Number of Clients}\label{sec:exp_client_num}
Figure~\ref{fig:client_num} compares FedCross with five baselines on the CIFAR-10 dataset using the ResNet-20 model within a non-IID scenario ($\beta=0.5$), where the total number of clients investigated in the subfigures is 50, 100, 200, 500, and 1,000, respectively. For each case, we selected 10\% of clients to participate in local training.
From Figure~\ref{fig:client_num}, we can observe that FedCross can achieve the best inference accuracy for all the cases.
Note that in this experiment, since the total number of samples is fixed, the larger the total number of clients, the smaller the amount of data assigned to each client.
As a result, we can find that when the number of clients increases, all the investigated FL methods need to use more training rounds for convergence.


\begin{table}[ht]
\vspace{-0.15in}
\centering
\caption{Test accuracy comparison with different $\alpha$ settings
}
\vspace{-0.1in}
\label{tab:criteria}
\scriptsize
\begin{tabular}{|c|c|c|c|}
\hline
\multirow{2}{*}{\bf $\alpha$} & \multicolumn{3}{c|}{\bf Selection Criteria} \\
\cline{2-4}
 & \bf In-Order & \bf Highest Similarity & \bf Lowest Similarity\\
\hline
$0.5$& $56.42\pm 0.54$& $56.33\pm 0.23$& ${\bf 56.81\pm 0.91}$ \\
$0.8$& $56.66\pm 0.46$& $55.83\pm 0.85$& ${\bf 57.78\pm 0.65}$ \\
$0.9$& ${\bf 58.69\pm 0.46}$& $46.91\pm 0.97$& $58.61\pm 0.48$ \\
$0.95$& $59.12\pm 0.62$& $49.94\pm 0.94$& ${\bf 59.47\pm 0.38}$ \\
$0.99$& $59.86\pm 0.40$& $49.70\pm 1.33$& ${\bf 62.16\pm 0.42}$ \\
$0.999$& $40.85\pm 1.82$& $32.51\pm 3.39$& ${\bf 46.83\pm 1.14}$ \\
\hline
\end{tabular}
	\vspace{-0.1in}
\end{table}

\subsection{Ablation Studies (RQ4)}

\subsubsection{Evaluation of Model Selection Strategies}\label{sec:exp_criteria}

Table~\ref{tab:criteria} presents the 
classification
performance using 
three model selection strategies on the CIFAR-10 dataset within
a non-IID scenario ($\beta=1.0$).
From Table~\ref{tab:criteria}, we can observe that {\it the lowest similarity} strategy can achieve the best performance for five out of the given 
six  $\alpha$ settings. 
Note that  {\it the highest similarity} strategy achieves
the worst performance for all the $\alpha$ settings.
This is because the {\it the highest similarity} strategy makes
middleware models with high similarity gradually get closer, while the models with low similarities become far away from each other, resulting in higher aggregation difficulty for the global model.
On the contrary, {\it the lowest similarity} reduces the distances between models with low similarities in each round of aggregation, which forces all the models to roughly optimize their local training towards similar directions.
Regarding 
 the in-order strategy, since every two models are
 aggregated within a  finite  number of
 rounds, the similarities  between models
 will be limited to a certain range. However, its efficiency will be relatively lower compared with the one achieved by the highest similarity strategy. In summary, we recommend using either the lowest similarity strategy or the in-order strategy to select the collaboration model.

\subsubsection{Evaluation of Aggregation Rate $\alpha$}\label{sec:exp_rate}
\begin{figure}[ht]
\vspace{-0.05 in}
	\centering
	\subfigure[In-Order]{
		\centering
		\includegraphics[width=0.18\textwidth]{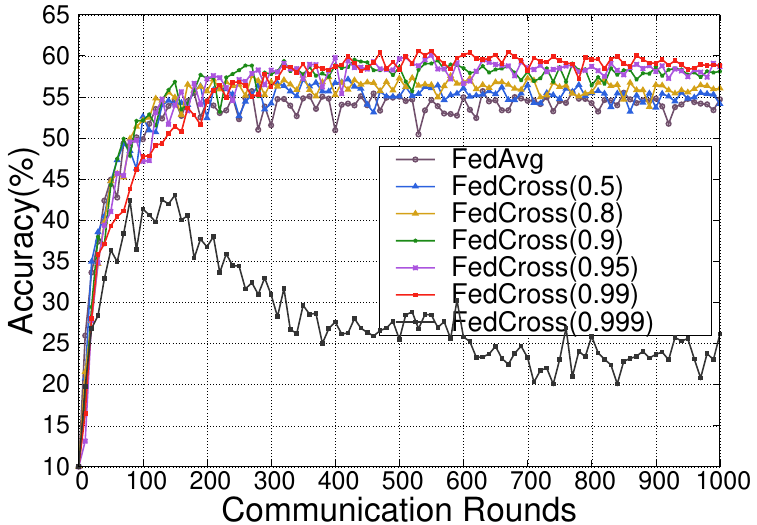}
		\label{fig:abl_io}
	}\vspace{-0.1 in}
	\subfigure[Lowest Similarity]{
		\centering
		\includegraphics[width=0.18\textwidth]{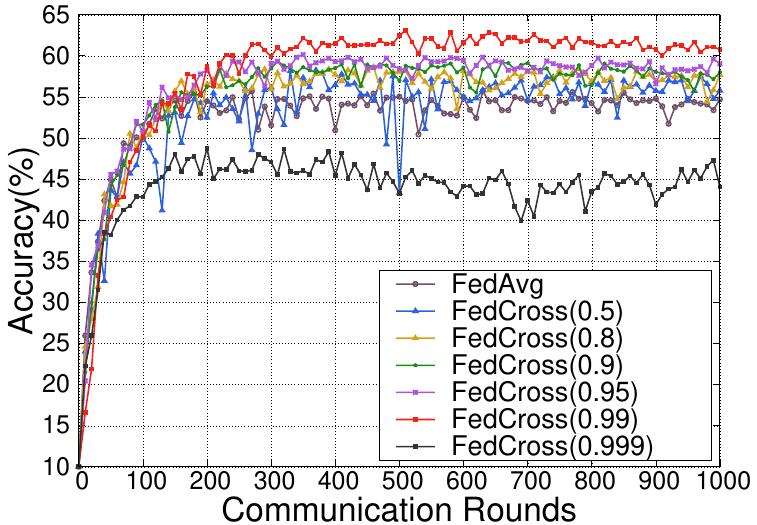}
		\label{fig:abl_ls}
	}
	\caption{Learning curves of CNN-based FedCross with  different  $\alpha$ settings
 within a non-IID scenario ($\beta=1.0$).}
	\label{fig:abl_alpha}
 	 \vspace{-0.05in}
\end{figure}

Figure~\ref{fig:abl_alpha} presents learning curves of both the in-order and lowest similarity strategies with six different settings of $\alpha$. In Figure~\ref{fig:abl_alpha}, FedCross performs best when $\alpha=0.99$.
We can observe that, as the value of $\alpha$ decreases, the performance of FedCross gradually decreases.
However, when $\alpha=0.999$, the performance of FedCross drops sharply. This is because the value of $\alpha$ is too large, which leads to less knowledge acquisition from the collaboration model.
In other words, reducing the distance between models in each round of aggregation cannot offset the increase in model distance in each round of training.
Therefore, the distances between models will gradually increase, resulting in a sharp decline in the performance of the global model.
From this figure, we can find that a large $\alpha$ will improve the performance of FedCross since it supports the model aggregation in a  more fine-grained way. 
Note that a  large $\alpha$  may cause a sharp performance drop for the global model.
In our experiments, FedCross achieves the best performance when $\alpha=0.99$. 
We recommend using a $\alpha=0.99$ in FedCross.

\subsubsection{Evaluation of Training  Acceleration Methods}\label{sec:acceleration}
\begin{figure}[h]
\vspace{-0.15in}
	\centering
	\subfigure[non-IID ($\beta=0.1$)]{
		\centering
		\includegraphics[width=0.18\textwidth]{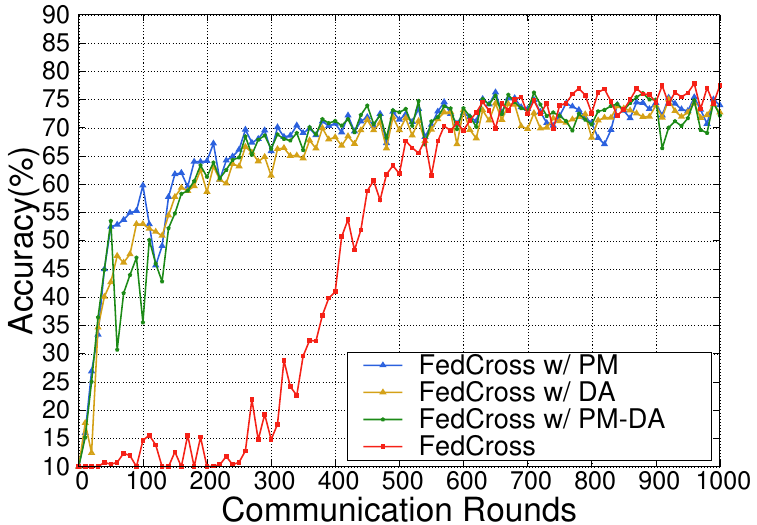}
		\label{fig:abl_ac_d0.1}
	}
	\subfigure[IID]{
		\centering
		\includegraphics[width=0.18\textwidth]{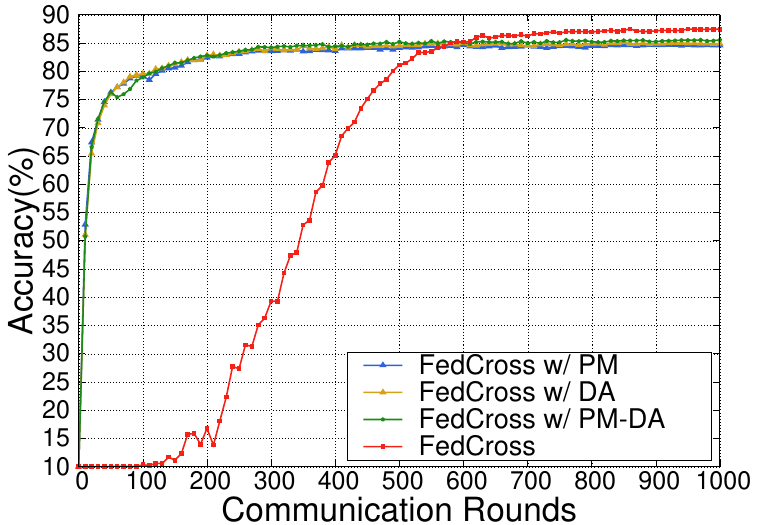}
		\label{fig:abl_ac_iid}
	}
 \vspace{-0.05in}
	\caption{Learning curves of VGG-16-based FedCross with different training acceleration methods  on CIFAR-10 dataset.}
	\label{fig:abl_ac}%
 \vspace{-0.15in}
\end{figure}

We evaluated the performance of two training acceleration methods on the CIFAR-10 dataset using the VGG-16 model.
Here, we considered three variants for FedCross.
The first variant ``FedCross w/ PM'' uses propeller models to speed up training in the first 100 FL rounds.
The second variant ``FedCross w/ DA'' uses dynamic $\alpha$ to speed up training for the first 100 FL rounds.
The third variant ``FedCross w/ PM-DA'' uses propeller models for the first 50 rounds and dynamic $\alpha$ for the
following 50 rounds to speed up training.
Figure~\ref{fig:abl_ac} presents the learning curves of FedCross in both
non-IID ($\beta=0.1$) and IID scenarios.
From Figure~\ref{fig:abl_ac}, we can find that all the variants can significantly accelerate the training, but will slightly reduce the models' accuracy. 
In the non-IID scenario, the performance of the three variants is similar. In the IID scenario, the performance of ``FedCross w/ PM-DA'' is higher than the other two variants.

\subsection{Discussion}
\subsubsection{Privacy Preserving}
Similar to 
 traditional one-to-multi FL methods, for FL training 
 FedCross does not need any data distribution information for each local client.
FedCross does not attempt to restore the user data by analyzing the model for each upload model.
For each dispatched model, since it is aggregated with a collaborative model, and the model is dispatched randomly,
clients cannot restore client data through the model and do not know the sources of received models.
In addition, since the model dispatching, local training, and model update processes of FedCross are the same as the ones of  FedAvg, FedCross can easily integrate existing
privacy-preserving techniques~\cite{triastcyn2019federated,wei2020federated,ijcai2021p217} that are suitable for FedAvg to avoid privacy leaks.

\subsubsection{Limitations}
Although FedCross can achieve  better performance than the baselines, its slow convergence  on complex models is still 
a severe limitation that is worthy of further study.
Although our proposed acceleration method can partially 
alleviate this problem, it may lead to slight performance degradation. 
Therefore, we need a more powerful acceleration method that does not affect the overall classification performance.
Furthermore, at present, we only considered  heterogeneous data for FedCross, where
FedCross cannot deal with the training of heterogeneous models. These will be an interesting topic for our future work.

\section{Conclusions}
\label{conclusion}



Due to the classic
FedAvg-based local model aggregation scheme, traditional Federated Learning (FL) methods greatly suffer from 
the problems of slow convergence as well as low classification accuracy, especially for non-IID scenarios. 
To address this problem, this paper presents a novel FL framework named FedCross, which adopts our proposed 
multiple-to-multiple training scheme, i.e., multi-model cross aggregation.
During the FL training, FedCross maintains a small set of intermediate models on the 
cloud server for the purpose of weighted fusion of similar local models. Since Fedcross 
fully 
respects the convergence characteristics of individual clients rather than  simply 
averaging their local models, the local models can quickly converge to their local 
optimum counterparts. Comprehensive experimental results on 
well-known datasets show that FedCross outperforms state-of-the-art FL methods significantly in both IID and non-IID scenarios without causing 
extra communication overhead.

\section*{Acknowledgments}
This research/project is supported by the National Research Foundation Singapore and DSO National Laboratories under the AI Singapore Programme (AISG Award No: AISG2-RP-2020-019), the Natural Science Foundation of China (62272170), the National Research Foundation, Singapore, and the Cyber Security Agency under its National Cybersecurity R\&D Programme (NCRP25-P04-TAICeN),
 ``Digital Silk Road'' Shanghai International Joint Lab of Trustworthy Intelligent Software (22510750100), and Natural Science Foundation (NSF CCF-2217104).
Any opinions, findings, conclusions, or recommendations expressed in this material are those of the author(s) and do not reflect the views of National Research Foundation, Singapore, and Cyber Security Agency of Singapore.
Mingsong Chen is the corresponding author (mschen@sei.ecnu.edu.cn).

\balance

\bibliographystyle{IEEEtran}
\bibliography{reference}


\end{document}